\documentclass[letterpaper]{article} % DO NOT CHANGE THIS
\usepackage[preprint]{aaai2027}  % DO NOT CHANGE THIS
\usepackage{graphicx} % DO NOT CHANGE THIS
\usepackage{natbib}  % DO NOT CHANGE THIS AND DO NOT ADD ANY OPTIONS TO IT
\usepackage{caption} % DO NOT CHANGE THIS AND DO NOT ADD ANY OPTIONS TO IT
\usepackage{algorithm}
\usepackage{algorithmic}

\usepackage{xcolor}
\usepackage[table]{xcolor}

\usepackage{booktabs} 
\usepackage{makecell}
\usepackage{tabularx}
\usepackage{array}
\usepackage{comment}
\newcolumntype{L}[1]{>{\raggedright\arraybackslash}p{#1}}

\newlength{\benchlabelwidth}
\newcommand{\exampleblock}[4]{%
  \begingroup
  \footnotesize
  \setlength{\tabcolsep}{0pt}%
  \renewcommand{\arraystretch}{1.08}%
  \begin{tabularx}{\hsize}{@{}
    >{\raggedright\arraybackslash\bfseries}p{3em}
    @{\hspace{0.35em}}
    >{\raggedright\arraybackslash}X
    @{}}
    Spec. & #1 \\
    Pos.  & #2 \\
    Gen.  & #3 \\
    Neg.  & #4 \\
  \end{tabularx}%
  \endgroup
}

\usepackage{colortbl}

\definecolor{scorebest}{RGB}{185,185,185}      % darker gray
\definecolor{scoresecond}{RGB}{225,225,225}    % lighter gray

\newcommand{\err}[1]{\textcolor{gray}{\,$\pm$\,#1}}
\newcommand{\best}[2]{%
  \cellcolor{scorebest}#1\err{#2}}
\newcommand{\second}[2]{%
  \cellcolor{scoresecond}#1\err{#2}}
\newcommand{\other}[2]{%
  #1\err{#2}}

\usepackage{newfloat}
\usepackage{listings}
\DeclareCaptionStyle{ruled}{labelfont=normalfont,labelsep=colon,strut=off} % DO NOT CHANGE THIS
\floatstyle{ruled}
\newfloat{listing}{tb}{lst}{}
\floatname{listing}{Listing}
\title{Localized Adaptation Reveals Distinct Learning Signatures in Transformers}
\author {
    Rebecca Ramnauth
    and Brian Scassellati
}
\affiliations {
    Yale University \\
    rebecca.ramnauth@yale.edu
}

\begin{document}

\maketitle

\begin{abstract}
Transformer adaptation is typically distributed across model depth, even when the intended change is narrow. We investigate how adaptation site shapes what a model learns, how well that learning generalizes, and how selectively it is applied. We introduce a controlled benchmark spanning five objectives (lexical binding, factual association, behavioral policy learning, causal mapping, and procedural reasoning) and define each objective's ``adaptation geometry'' as its profile of acquisition, transfer, and boundedness under full-stack and early-, middle-, or late-layer LoRA. The objectives exhibit distinct geometries. Lexical binding favors early-layer adaptation for acquisition and boundedness but requires broader updates for transfer; factual association favors later layers among localized adapters; behavioral learning separates late-layer action acquisition from middle-layer policy gating; and causal and procedural transfer benefit most from middle- or full-stack adaptation. These patterns largely persist under parameter-matched controls, and most corresponding directional contrasts replicate across five model families. \textcolor{black}{These findings establish adaptation site as a key design variable for controlling what models learn, generalize, and leave unchanged.}  %for improving learning transfer and boundedness while limiting unintended change. %rather than an arbitrary engineering choice.

%Transformer adaptation is typically applied broadly across model depth, even when the intended change is narrow in scope (e.g., learning a new fact or lexical mapping). This can entangle unrelated capabilities and obscure where different forms of learning are best supported inside the model. We study localized adaptation as a diagnostic tool for characterizing learning-type-specific adaptation geometry. We construct a synthetic benchmark in which each task instance is generated from a latent specification, allowing us to compare distinct learning objectives (lexical binding, factual association, causal mapping, behavioral policy learning, and procedural reasoning) under matched adaptation conditions. We then evaluate how adaptation site affects acquisition, transfer, and boundedness, and use these localization profiles to characterize objective-specific adaptation geometry. By comparing full-stack adaptation with localized updates across depth and module type, we define localization profiles for each learning objective. This framework asks whether different kinds of adaptation succeed and fail in systematically different regions of transformer models, offering a path toward more interpretable, bounded, and reversible model adaptation.
\end{abstract}

%We then evaluate how adaptation site affects target acquisition, generalization, negative-control behavior, interference, reversibility, and representational propagation. 

% Uncomment the following to link to your code, datasets, an extended version or similar.
% You must keep this block between (not within) the abstract and the main body of the paper.
\begin{links}
    \link{Code}{github.com/rramnauth2220/adaptation-geometries}
     %\link{Datasets}{https://aaai.org/example/datasets}
     %\link{Extended version}{https://aaai.org/example/extended-version}
\end{links}

\section{Introduction}
Modern language models are increasingly adapted after pretraining. They are fine-tuned for new domains, edited to update factual knowledge, aligned toward preferred behaviors, and extended through parameter-efficient adapters. Still, adaptation is largely treated as a globally distributed operation. Even when the intended change is narrow, such as learning a new fact or behavioral rule, updates are applied across large portions of the network without a clear account of where different forms of learning are best supported. This raises the concern that narrowly intended updates may become entangled with capabilities that were not meant to change. %Even when the intended change is narrow, such as learning a new fact or behavioral rule, updates are commonly applied across large portions of the transformer stack without a clear account of where different forms of learning are best supported. This has produced highly capable systems, but it also raises a concern that updates due to certain forms of learning can become entangled when only one capability is intended to change. 

\textcolor{black}{Adaptation ``success'' is not just a question of whether the target behavior is learned. A useful update should also transfer to appropriate held-out contexts and remain bounded to the contexts in which it should apply. These properties can diverge. For example, a model may acquire a new fact but over-apply it to related false cases; it may learn a behavioral rule but fail to withhold it when the rule is irrelevant. If different regions of the transformer support these components differently, then adaptation site becomes a practical design choice rather than a mere implementation detail.}

At the same time, a growing body of interpretability and probing work suggests that transformer representations are not functionally homogeneous across depth. Early layers are closely tied to lexical and syntactic processing, middle layers with increasingly abstract or relational information, and later layers with task execution and output behavior.\footnote{This description is intended as a directional characterization rather than a strict modular claim. Modern transformers reuse and transform information across depth, so any localization hypothesis should be treated probabilistically.} 

Accordingly, one may hypothesize that some forms of learning may only require localized interventions at the layer subnetwork where the relevant abstract naturally lives. From this, one may put forth that indiscriminately distributing updates across the network may be unnecessarily invasive. We therefore study the \emph{adaptation geometry} of distinct learning objectives. Rather than treating adaptation success as a single aggregate outcome, we ask how different regions of the transformer support different components of learning. For instance, we evaluate whether a target behavior is acquired, whether it transfers to held-out contexts, and whether it remains bounded to the contexts in which it should apply. The central hypothesis is that adaptation quality depends on whether the adaptation site matches the functional character of the learning objective itself.

\textcolor{black}{This perspective differs from work on parameter-efficient fine-tuning and layer-wise ablation, which often emphasizes computational efficiency or sensitivity. Our goal is instead to characterize the dynamic between learning type and adaptation locus. If distinct forms of learning can be reliably localized, models may acquire new knowledge and behavior with less disruption to existing capabilities, reduced interference, and more interpretable control over the scope of adaptation.} %This stance differs from work on parameter-efficient fine-tuning or layer-wise ablations, which primarily focus on computational efficiency or sensitivity analysis. Our goal is instead to characterize the relationship between learning type and adaptation locus. In doing so, we frame localization as a potential mechanism for more targets and scope-aware model adaptation. If certain forms of learning can be reliably isolated to specific regions of the network, this could enable systems to acquire new behaviors without globally perturbing existing capabilities, reduce catastrophic interference, and support more interpretable forms of behavioral control. 

%More broadly, the work reframes transformer adaptation as a problem of representational compatibility: identifying where in a model a given kind of learning most naturally belongs.

\section{Background}
%The literature increasingly supports the idea that transformer representations are not functionally homogeneous across depth. 
Prior work provides several reasons to expect depth-wise structure in transformer representations. Probing studies suggest that different kinds of linguistic information become recoverable at different depths; earlier layers tend to capture more local lexical and syntactic features, while later layers increasingly reflect higher-level semantic and task-relevant information \citep{tenney2019bert}. Related work on knowledge neurons similarly suggests that some factual associations can be traced to identifiable units or substructures within pretrained transformers \citep{dai2022knowledge}.

%Prior work provides several reasons to expect depth-wise structure in transformer representations. For instance, probing studies have shown that linguistic information often emerges in a partially ordered manner, with earlier layers more associated with local lexical and syntactic features and later layers increasingly encoding higher-level semantic and task-relevant information \citep{tenney2019bert}. Related work on knowledge neurons further suggests that some factual associations can be traced to identifiable internal units or substructures within pretrained transformers \citep{dai2022knowledge}.

The idea that learning is localizable is well known. From this foundation, we move from representational diagnosis to \textit{adaptation geometry} by asking where new adaptations should occur when the model is given different learning objectives. In this section, we review the related work that contributes to our intuition that such adaptation geometries may exist, can be distinguished under different learning conditions, and has implications for adaptation performance.

\subsection{Parameter-Efficient Adaptation}
Parameter-efficient fine-tuning methods adapt large pretrained models by updating only a small subset of parameters, making them a natural setting for studying where adaptation occurs within a transformer. \textcolor{black}{Low-rank adaptation (LoRA)}, for instance, freezes the pretrained weights and introduces low-rank trainable updates into transformer modules, substantially reducing the number of trainable parameters while preserving downstream performance \cite{hu2022lora}. Adapter-based methods pursue a similar goal by inserting small trainable modules into pretrained transformers, enabling task-specific adaptation without full-model fine-tuning. Importantly, work has shown that adapter placement and capacity need not be uniform across the transformer stack. For example, AdapterDrop removes adapters from lower layers to improve efficiency with limited performance loss \citep{ruckle2021adapterdrop}, while AdaLoRA dynamically allocates low-rank adaptation budget across weight matrices according to their estimated importance \citep{zhang2023adalora}. 

These methods show that trainable capacity need not be distributed uniformly across a transformer. However, in much of this work, localization is treated primarily as an engineering choice (reducing computational costs while preserving downstream task performance). Layer selection and rank allocation are therefore typically evaluated by how efficiently they approximate full fine-tuning under a constrained parameter budget. Building on this observation, we ask whether the location of a successful adaptation is itself informative. What does the success or failure of a localized update reveal about the kind of learning being introduced? %Building on this observation, we ask a different question. If adaptation can succeed when restricted to only part of the network, then its location may itself be informative. Less well understood is what the success or failure of a localized update reveals about the kind of learning being introduced.

\subsection{Model Editing and Localized Knowledge Updates}

Model editing aims to modify specific model behaviors or factual associations without retraining the entire model. ROME introduced a causal tracing approach for locating factual associations in GPT-style models and showed that rank-one updates to mid-layer feed-forward modules can edit individual factual memories \citep{meng2022locating}. MEMIT extended this approach to mass-editing, enabling many factual associations to be inserted or modified simultaneously \citep{meng2022mass}. This line of work provides strong evidence that at least some forms of knowledge can be edited through localized interventions in the transformer.

However, most model-editing work centers on declarative factual associations because it is the most operationally tractable case: it has a clear unit of intervention, typically a subject--relation--object association, a target output, and established metrics \cite{wang2024knowledge}. This leaves open whether the localization patterns observed for factual recall reflect a general principle of transformer adaptation or a structure specific to the factual-editing setting. Broadening the scope to compare multiple qualitatively distinct learning types allows us to answer this and also explore how different forms of learning produce different adaptation signatures across the transformer stack. 

%This broader comparison is harder because ``learning type'' is not as easy to operationalize. A factual update can be tested with “The Eiffel Tower is in...” But a causal update requires checking whether the model propagates consequences under interventions; a behavioral update requires measuring response tendencies across contexts; procedural reasoning requires looking at multi-step computation rather than a single recalled answer. Those are not just different datasets---they may require different evaluation logic.
%A factual edit asks the model to retrieve a new relation, whereas a causal update asks it to propagate consequences under interventions or procedural reasoning asks it to change the trajectory of intermediate computation. Although all may involve ``associative learning'' in the broad sense, they should not be expected to localize or fail in the same way. 

\subsection{From Locality to Adaptation Geometry}
Model editing aims to make targeted changes while preserving unrelated behavior, a property known as \textit{locality}. This goal is typically formalized through three metrics: an edit should produce the intended answer (acquisition), generalize across equivalent prompt formulations (transfer), and avoid applying where it should not (boundedness) \cite{de2021editing, mitchell2021fast, meng2022locating}. Memory-based and semi-parametric editors further emphasize controlling the scope of an edit, often by deciding when an edit should apply and when the base model should remain unchanged \cite{mitchell2022memory}. 

Recent work shows that this notion of locality is necessary but insufficient. Even when individual edits appear successful, repeated or large-scale editing can degrade previously edited facts, disrupt broader knowledge structure, weaken safety-relevant behavior, or trigger broader model collapse \cite{gupta2024model, gu2024model, yang2024butterfly, li2024should}. These results motivate concern about the \textit{scope} of adaptation: updates must produce the intended behavior while remaining bounded to the contexts in which that behavior is appropriate.

We build on this concern, but shift the question from whether a single edit is local to how the location of adaptation shapes the kind of learning that occurs. In model editing, locality is usually evaluated after an update has been made---did the intended change occur, and did it remain appropriately scoped? We instead ask whether those outcomes depend systematically on where the update is placed in the transformer. If the same objective is adapted in early, middle, late, or full-stack regions, does it produce the same balance of acquisition, transfer, and boundedness, or does each location induce a different pattern of success and failure?

We call this pattern an objective's \emph{adaptation geometry}, i.e., the profile of acquisition, transfer, and boundedness that emerges when a learning objective is constrained to different regions of the model. This framing lets us compare whether different objectives rely on different adaptation loci and fail in different ways when placed in poorly matched regions.

\section{Methodology}

\label{sec:methodology}

%We study whether different learning objectives exhibit distinct localization profiles within language models. Rather than asking only whether a model can learn a target behavior, we ask how the site of adaptation affects three outcomes: whether the target behavior is acquired, whether it transfers to held-out contexts, and whether it remains bounded to the contexts where it should apply. 

We use localized LoRA to characterize the adaptation geometry of various learning objectives in language models.

\subsection{Benchmark Construction}
We construct a benchmark spanning five operationally distinct learning objectives: lexical binding, factual association, behavioral policy, causal mapping, and procedural reasoning. These objectives are not exhaustive or ontologically separate, but differ in the structure the model must acquire and generalize \cite{jacobs2021measurement, raji2021ai}.

Each task instance is generated from a latent specification, which serves as the unit of learning. For lexical binding, a latent specification maps a novel surface form to an existing concept. For factual association, it defines a subject--relation--object association. For behavioral policy learning, it defines a context condition under which a response action should be applied. For causal mapping, it defines a synthetic causal rule linking an intervention or activating event to an effect through an underlying mechanism. For procedural reasoning, it defines a procedure with required setup steps, intermediate actions, and a completion condition. 

From each latent specification, we generate training and evaluation examples. The evaluation split contains in-distribution examples, paraphrase examples, generalization examples, and negative-control examples. In-distribution and paraphrase examples test whether the target behavior has been acquired under unseen formulations of the same latent specification. Generalization examples test whether the learned behavior transfers beyond the exact adaptation pattern. Negative-control examples are superficially similar prompts for which the learned behavior should not apply, allowing us to measure whether adaptation remains bounded rather than overgeneralizing.

For each learning objective, we construct 25 latent specifications. Each specification contributes 12 training examples and 22 evaluation examples: 5 in-distribution examples, 5 paraphrase examples, 6 generalization examples, and 6 negative-control examples. Thus, for each objective, budget, and seed, in-distribution and paraphrase evaluations contain \(n=125\) examples each, while generalization and negative-control evaluations contain \(n=150\) examples each.

\subsection{Calibration and Budget Selection}
\label{sec:calibration-method}

Equal dataset size does not imply equal adaptation difficulty. A lexical binding may be learned from relatively few examples, whereas a behavioral policy may require many more examples to define when the learned behavior should and should not apply. To avoid confounding localization with task difficulty, we calibrate the training budget separately for each objective before running localization experiments.

A training budget denotes the number of training examples per latent specification. Calibration proceeds in two stages. First, we run a ``coarse'' full-stack LoRA sweep over candidate budgets to identify the transition point in which each objective becomes learnable. Second, we run neighboring candidate budgets across multiple seeds near that transition point. For each objective, we select the smallest candidate budget whose mean acquisition and transfer are within 95\% of the best observed mean. We treat boundedness as a diagnostic outcome rather than a hard selection threshold, since boundedness can reveal overgeneralization even when acquisition and transfer are high.

This procedure selects budget 10 for lexical binding, behavioral policy, and causal mapping, and budget 8 for factual association and procedural reasoning (Appendix B). We hold these budgets fixed across the localization experiments.

\subsection{Evaluation Metrics}
\label{sec:evaluation-metrics}

We evaluate each adapted model along three core metrics: acquisition, transfer, and boundedness. Acquisition measures whether the target behavior is learned under direct and paraphrased formulations of the same latent specification. Let $\mathrm{Acc}_{x}(T,s)$ denote accuracy on evaluation split $x$ for objective $T$ and adaptation site $s$. We define acquisition as:
\[
A_T(s) =
\frac{1}{2}
\left(
\mathrm{Acc}_{\mathrm{id}}(T,s)
+
\mathrm{Acc}_{\mathrm{para}}(T,s)
\right).
\]
Transfer measures generalization to held-out examples from the same latent specification, $R_T(s) = \mathrm{Acc}_{\mathrm{gen}}(T,s)$. Boundedness measures whether learned behavior is withheld in negative-control contexts,  
$B_T(s) = \mathrm{Acc}_{\mathrm{neg}}(T,s)$. We define an objective's \emph{adaptation geometry} as the profile of acquisition, transfer, and boundedness at each site:
\[
G_T(s) = \big(A_T(s), R_T(s), B_T(s)\big).
\]

Because negative controls differ by objective, boundedness is scored by objective. Lexical binding and factual association use strict negative-control accuracy. Behavioral learning uses policy-withholding accuracy, testing whether the model refrains from applying a learned policy when no policy action is warranted. Causal mapping uses null-effect accuracy, testing whether the model correctly predicts that the learned causal effect should not occur. Procedural reasoning uses incomplete-procedure accuracy, testing whether the model rejects partial, reordered, or invalid procedures.

\subsection{Localized Adaptation Regimes}
\label{sec:localized-adaptation-regimes}

For each calibrated objective, we train LoRA adapters under four regimes: full-stack, early-layer, middle-layer, and late-layer adaptation. Full-stack adaptation updates adapters across all layers, while localized regimes restrict updates to one contiguous depth region with the base model frozen.

Experiments are per-layer matched: each adapted layer uses the same LoRA rank across regimes. Thus, localized comparisons isolate the effect of adaptation location under fixed per-layer capacity. Because full-stack adaptation updates more layers and has greater total capacity, we treat it as an empirical upper bound and use it as a reference for interpreting localized adaptation.

\subsection{Mislocation Analysis}
\label{sec:mislocation-method}

Localization also allow us to examine the consequences of adapting an objective in a region that is poorly matched to its observed adaptation geometry. After identifying the strongest localized region for a learning objective, we compare it against a non-preferred (i.e., mislocated) region.

We define the mislocation penalty as the performance difference between an objective's preferred localized site and a non-preferred site. Positive values indicate that respecting the objective's adaptation geometry improves performance. 

\begin{comment}
For objective \(T\), metric \(m\), preferred site \(s^+\), and mislocated site \(s^{-}\), we define a mislocation penalty as:
\[
\Delta_{\mathrm{misloc}}(T,m)
=
m_T(s^+) - m_T(s^{-}).
\]
Positive values indicate that respecting the objective's adaptation geometry improves performance. %This is computed from the same localized adaptation runs; it is not a separate experimental design. Its purpose is to characterize what fails when adaptation is constrained to a region that is poorly aligned with the objective.
\end{comment}

\subsection{Statistical Analysis}
\label{sec:statistical-analysis}

We analyze localization effects using seed-level paired contrasts with bootstrap confidence intervals. Rather than exhaustively testing all pairwise condition differences, we define planned contrasts corresponding to each objective's hypothesized localization profile (Appendix A). %For example, lexical binding contrasts early-layer adaptation against middle- and late-layer adaptation, factual association contrasts late-layer adaptation against early-layer adaptation, and causal mapping contrasts middle-layer adaptation against early- and late-layer adaptation.

To test the broader claim that learning objectives exhibit distinct geometries, we perform an omnibus permutation test over seed-level adaptation-geometry profiles. The observed statistic is the between-objective dispersion of \(\Delta G_T\) profiles. We then permute objective labels across seed-level profiles and recompute this statistic. This tests whether objective labels explain localization geometry beyond what would be expected from the observed profile variability alone.

\subsection{Parameter-Matched Controls}
\label{sec:parameter-matched-method}
Experiments are per-layer matched, so full-stack adapters have more trainable parameters than localized adapters. To check whether localization profiles reflect adaptation site rather than parameter count alone, we include two total-parameter-matched controls. In the localized-expanded-rank control, localized windows receive proportionally larger rank to match the full-stack layer-rank product \((32 \times 8 = 8 \times 32)\). In the full-reduced-rank control, full-stack adaptation receives proportionally smaller rank \((32 \times 2 = 8 \times 8)\). These controls test whether the qualitative localization signatures persist when approximate adapter capacity is matched.

 %The main localization experiments use a per-layer-matched design: full-stack, early-layer, middle-layer, and late-layer adapters use the same LoRA rank per adapted layer. This means that full-stack adaptation has more total trainable parameters than a localized window and should be interpreted as a calibrated upper-bound condition rather than a capacity-matched comparison.

%For Llama-3.1-8B-Instruct \cite{grattafiori2024llama}, which has 32 transformer layers, we define three localized regions using contiguous 8-layer windows: early layers, middle layers, and late layers. Full-stack adaptation inserts LoRA adapters across all layers. The localized conditions therefore allow us to test what forms of learning remain possible when updates are constrained to different regions of model depth.

\subsection{Cross-Model Robustness Analysis}
\label{sec:cross-model-method}
The main experiments use Llama-3.1-8B \cite{grattafiori2024llama} as the base model, with base parameters frozen and LoRA updates inserted into transformer projection modules. We estimate the initial adaptation geometries using objective-specific budgets calibrated on this model.

We then test whether these profiles recur across additional models, using Mistral-7B \cite{jiang2023mistral7b}, Gemma-2-9B \cite{team2024gemma}, OLMo-2-7B \cite{olmo20242}, and Qwen2.5-14B \cite{qwen2}. We selected these models because they are open-weight, instruction-tuned transformers at a scale where multi-objective localized LoRA adaptation remains computationally feasible, while still spanning distinct model families and architectural/training pipelines. For this cross-model analysis, we transfer the Llama-calibrated objective budgets rather than recalibrating every model--objective pair. The analysis is therefore a robustness test of the Llama-calibrated protocol, not a fully optimized comparison across models.

Because models differ in depth, we define localized regions by normalized layer position rather than fixed layer indices. We use quarter-depth windows rather than thirds to keep each localized condition narrowly constrained while still probing distinctively early, middle, and late regions of the network. These regions serve as representative adaptation sites, not an exhaustive partition of all layers.

For each model, objective, seed, and site, we compute the adaptation geometry based on acquisition, transfer, and boundedness. We also report robustness diagnostics, including best-region agreement across models, similarity of localization profiles, and replication of planned mislocation penalties. When full-stack performance is weak under a transferred budget, we interpret the corresponding profile as a possible model--budget mismatch rather than as definitive evidence about that model's adaptation geometry.

\section{Results of Localized Adaptation} \label{sec:localization}
Having selected a calibration budget for each learning objective (Appendix B), we next ask whether different forms of learning exhibit distinct depth-wise adaptation signatures. Because full-stack LoRA updates more layers than each localized window, we treat it as an empirical upper-bound condition and use the localized conditions to diagnose which regions support different components of learning. 

\subsection{Cross-Objective Overview}
\begin{figure*}[t]
    \centering
    \includegraphics[width=\linewidth]{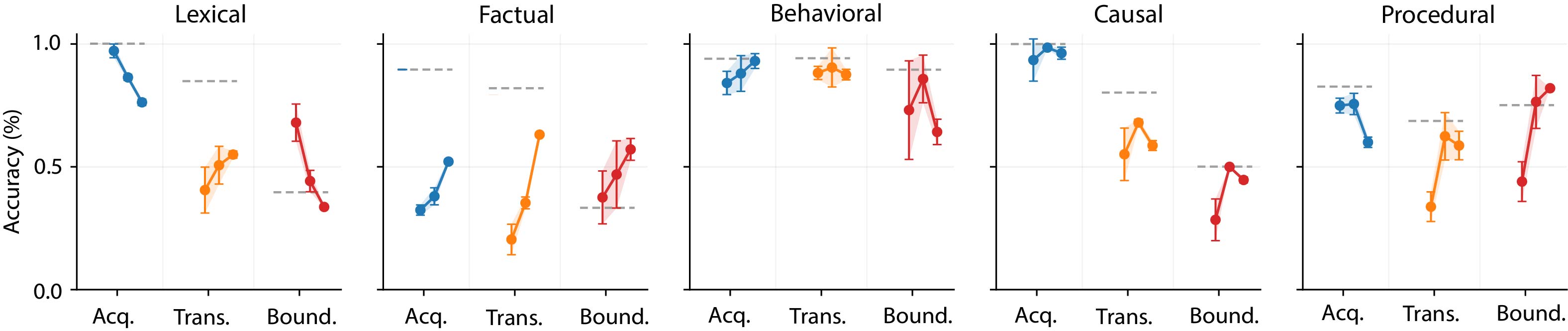}
    \caption{Cross-objective geometries for five learning objectives in Llama-3.1-8B. \textcolor{black}{Within each metric, connected points show seed means across early, middle, and late adaptation; error bars indicate standard deviations, and dashed lines denote full-stack means. Signatures vary by objective: lexical binding localizes early, factual association favors later layers, behavioral policy is distributed, causal transfer favors middle/full-stack learning, and procedural reasoning trades middle-layer transfer for late-layer boundedness. In many cases, targeted learning outperforms the full-stack.} Full statistics appear in Appendix C, Table \ref{tab:cross-objective-localization}.}

    %\caption{
    %Cross-objective localization signatures for five learning objectives in Llama-3.1-8B. Bars show seed means, error bars show standard deviations, and points show individual runs. Signatures vary by objective: lexical binding localizes early, factual association favors later layers, behavioral policy is distributed, causal transfer favors middle/full-stack adaptation, and procedural reasoning trades middle-layer transfer for late-layer boundedness. Full statistics appear in Appendix C, Table \ref{tab:cross-objective-localization}.}
    \label{fig:cross-objective-localization}
\end{figure*}

Fig.~\ref{fig:cross-objective-localization} summarizes localization profiles across all five calibrated objectives for Llama-3.1-8B. The results show that localization does not produce a single universal depth preference. Instead, each learning objective exhibits a distinct pattern across acquisition, transfer, and boundedness.

Lexical binding is the clearest early-localized objective. Early-layer adaptation nearly preserves acquisition while improving boundedness relative to full-stack adaptation, but it transfers poorly. Factual association shows the opposite pattern in that early-layer adaptation performs poorly, while late-layer is the strongest localized condition. Behavioral learning is more distributed, with late layers supporting acquisition and middle layers supporting bounded policy application. Causal mapping is strongest in the middle region among localized adapters, although full-stack adaptation remains best for transfer. Procedural reasoning also favors middle-layer adaptation for transfer, while late layers produce higher boundedness at the cost of weaker acquisition. 

Across parameter-matched controls, the main localization signatures are largely preserved (Appendix C, Table \ref{tab:parameter-matched-controls-app}). These patterns show that objectives differ not only in how well they can be adapted, but also in where acquisition, transfer, and boundedness are supported within the model. 

%We approach our results by recognizing that full-stack adaptation is the performance ceiling; localized adaptation reveals what kind of learning is happening and what tradeoffs full-stack adaptation hides. Wherein full-stack adaptation as a dedicated experimental condition tells us what works best, this section conveys that forced localization tells us why, where, and at what behavioral cost. 

\subsection{Lexical Binding is Early-Localizable}
\label{sec:localization-lexical}

%Lexical binding produced the clearest early-layer localization profile. 
For lexical binding, early adaptation nearly matched full-stack acquisition ($97.2\%$ vs. $99.7\%$) and improved boundedness ($68.0\%$ vs. $39.7\%$), while full-stack adaptation transferred learning best ($84.9\%$). Planned contrasts confirmed this early-layer signature. Early adaptation outperformed late and middle adaptation on acquisition by $20.9$ percentage points (pp; CI $[17.2,23.6]$)\footnote{All reported intervals are 95\% bootstrap confidence intervals.} and $10.8$ pp (CI $[7.6,13.2]$), respectively, and outperformed full-stack adaptation on boundedness by $28.3$ pp (CI $[25.0,30.7]$).

However, full-stack adaptation outperformed early adaptation on transfer by $44.3$ pp (CI $[37.3,48.0]$). Thus, early layers can acquire a novel lexical mapping while keeping it comparatively bounded, but robust generalization requires broader adaptation. %Parameter-matched controls preserved the same qualitative pattern, suggesting that the early lexical-binding signature is not explained by trainable parameter count alone.

\subsection{Factual Association Favors Later Adaptation}
\label{sec:localization-factual}

Factual association showed a later localization profile than lexical binding. Full-stack adaptation performed best overall ($89.6\%$ acquisition; $82.0\%$ transfer), while early adaptation performed poorly ($32.4\%$; $20.4\%$). Among localized conditions, late adaptation was strongest ($52.1\%$; $63.1\%$).

Late adaptation outperformed early adaptation on acquisition by $19.7$ pp (CI $[18.0,21.6]$) and transfer by $42.7$ pp (CI $[35.3,47.3]$). Full-stack nevertheless exceeded late adaptation on transfer by $18.9$ pp (CI $[16.7,22.0]$), whereas late adaptation improved boundedness by $23.8$ pp (CI $[22.0,25.3]$). Thus, unlike lexical binding, factual association favors later or more distributed updates, with localized adaptation trading transfer for boundedness.

\subsection{Behavioral Policy Learning is Distributed}
\label{sec:localization-behavioral}

Behavioral policy learning showed a distributed localization profile. Full-stack adaptation performed best overall ($94.0\%$ acquisition, $94.2\%$ transfer, and $89.6\%$ boundedness). Among localized conditions, late adaptation nearly preserved acquisition ($93.1\%$), whereas middle adaptation achieved greater transfer ($90.4\%$) and boundedness ($85.8\%$).

Late- exceeded middle-layer acquisition by $5.1$ pp (CI $[0.8,10.4]$), but middle adaptation improved boundedness by $21.6$ pp (CI $[15.3,26.7]$). Their transfer difference was small and uncertain ($2.9$ pp, CI $[-5.3,8.0]$). Thus, policy acquisition and gating appear partially separable. Late layers better support learning the action, whereas middle layers better preserve when it should be applied or withheld.\footnote{Here, boundedness could be inflated if the model defaulted to the withholding label \textsc{no \_policy \_trigger}. This occurred in $291/4800$ positive examples ($6.1\%$), indicating that boundedness was not explained by global collapse to policy withholding, though localized adapters can still differ in how conservatively they apply learned policies (detailed in Appendix C, Table \ref{tab:null-label-audits}).}

\subsection{Causal Mapping Requires Middle or Full-Stack Adaptation for Transfer}
\label{sec:localization-causal}

Causal mapping showed a middle-layer localization profile. Full-stack adaptation achieved saturated acquisition ($100.0\%$) and the strongest overall transfer ($80.2\%$). Among localized conditions, middle adaptation performed best in that it nearly preserved acquisition ($98.5\%$), matched full-stack boundedness ($50.0\%$), and achieved the strongest localized transfer ($68.0\%$).

Middle adaptation exceeded early and late adaptation on transfer by $12.9$ pp (CI $[2.0,22.0]$) and $9.3$ pp (CI $[8.7,10.0]$), respectively, and improved boundedness\footnote{Here, boundedness could be inflated if the model defaulted to the null label \textsc{no\_causal\_effect}. We therefore audited positive causal examples and found zero null-label predictions across all runs: $0/1500$ in-distribution, $0/1500$ paraphrase, and $0/1800$ generalization examples. Thus, causal boundedness is not explained by conservative collapse to the null response.} over early adaptation by $21.6$ pp (CI $[16.7,31.3]$). Full-stack nevertheless surpassed middle adaptation on transfer by $12.2$ pp (CI $[2.0,17.3]$). Thus, causal mapping favors middle-layer updates among localized adapters, but robust generalization still benefits from full-stack adaptation.

%Here, boundedness is measured by accuracy on negative controls whose correct label is \textsc{no\_causal\_effect}. This metric could be inflated if the model simply defaulted to this null label. Therefore, we audited positive examples and counted null-label predictions on examples whose correct answer was a non-null causal effect. Across all runs, the model predicted a null-effect on $0/1500$ in-distribution examples, $0/1500$ paraphrase examples, and $0/1800$ generalization examples. Thus, causal boundedness is not explained by conservative collapse to the null response.

\subsection{Procedural Reasoning Shows a Transfer--Boundedness Tradeoff}
\label{sec:localization-procedural}

Procedural reasoning showed a middle-layer localization profile among localized adapters. Full-stack adaptation performed best overall ($82.7\%$ acquisition; $68.7\%$ transfer), while middle adaptation provided the strongest localized balance ($75.6\%$ acquisition, $62.4\%$ transfer, and $76.4\%$ boundedness). Early adaptation transferred poorly ($33.8\%$), whereas late adaptation maximized boundedness ($82.0\%$) but reduced acquisition ($60.0\%$).

Middle adaptation exceeded early adaptation on transfer by $28.7$ pp (CI $[23.3,34.0]$) and exceeded late adaptation on acquisition by $15.6$ pp (CI $[12.0,21.2]$). Differences between full and middle transfer ($6.2$ pp, CI $[-1.3,17.3]$) and between late and middle boundedness ($5.6$ pp, CI $[-1.3,18.0]$) did not reliably differ from zero. Thus, middle adaptation offers the strongest localized balance across acquisition, transfer, and boundedness, while late-layer adaptation appears more conservative in that it rejects incomplete procedures well, but does so at the cost of weaker acquisition.\footnote{We audited conservative rejection by counting \textsc{procedure\_incomplete} predictions on positive examples. Across runs, this occurred on $138/4800$ examples ($2.9\%$), with the highest rate under late-layer adaptation ($100/1200 = 8.3\%$; Appendix C, Table \ref{tab:null-label-audits}). This suggests that late-layer boundedness partly reflects conservative rejection, though not global collapse.}

\subsection{Consequences of Mislocalized Adaptation}
\label{sec:mislocation-effects}
Mislocation penalties show that poorly placed adapters do not fail uniformly (Appendix C, Table \ref{tab:appendix-primary-mislocation-penalties}). Factual association suffers primarily transfer failure when forced into early layers. Conversely, lexical binding favors early layers for acquisition and boundedness but later layers for transfer. Behavioral policy separates action acquisition, favored by late adaptation, from policy gating, favored by middle adaptation. Causal and procedural learning both penalize early adaptation most strongly on transfer and boundedness. 

These results show that mislocation impacts how learning fails. Adapters placed in poorly matched regions may acquire labels without transferring them, preserve action selection without learning when to withhold it, or appear bounded only by becoming overly conservative. %Adaptation site should therefore be treated as a substantive design variable.

\subsection{Further Evidence for Distinct Geometries}
\label{sec:statistical-geometries}

The above contrasts test specific localization hypotheses for each objective. We additionally asked whether the overall geometry profiles differ across objectives. This analysis treats each objective's localized-minus-full profile across acquisition, transfer, and boundedness as a single geometry vector, and tests whether those vectors are more objective-specific than expected if objective labels were exchangeable.

An omnibus permutation test found significant heterogeneity across objectives (observed heterogeneity \(=0.238\), permutation \(p<.001\); $10{,}000$ permutations). This provides a diagnostic complement to the planned contrasts. Whereas the contrasts identify which components of learning fail under mislocation, the omnibus test evaluates the full adaptation-geometry profile and shows that these profiles are not interchangeable across objectives.%Thus, the evidence for adaptation geometry does not depend only on isolated pairwise comparisons; it also appears in the structure of the full localization profiles.

\section{Cross-Model Robustness of Geometries}
The primary experiments establish adaptation geometries in Llama-3.1-8B. We next test whether their directional signatures recur across model families using the same objective-specific budgets selected on Llama, without independently recalibrating each model. Because transferred budgets may underfit some model--objective pairs, we interpret cross-model localization profiles in light of the full-stack diagnostics shown in Appendix~D, Fig.~\ref{fig:cross-model-full-stack-performance}. Low full-stack acquisition or transfer indicates possible model--budget mismatch rather than strong evidence about geometry. Accordingly, we assess whether the primary \emph{directional} contrasts replicate, not whether every model yields the same best localized region.

\begin{figure}[t]
    \centering
    \includegraphics[width=\linewidth]{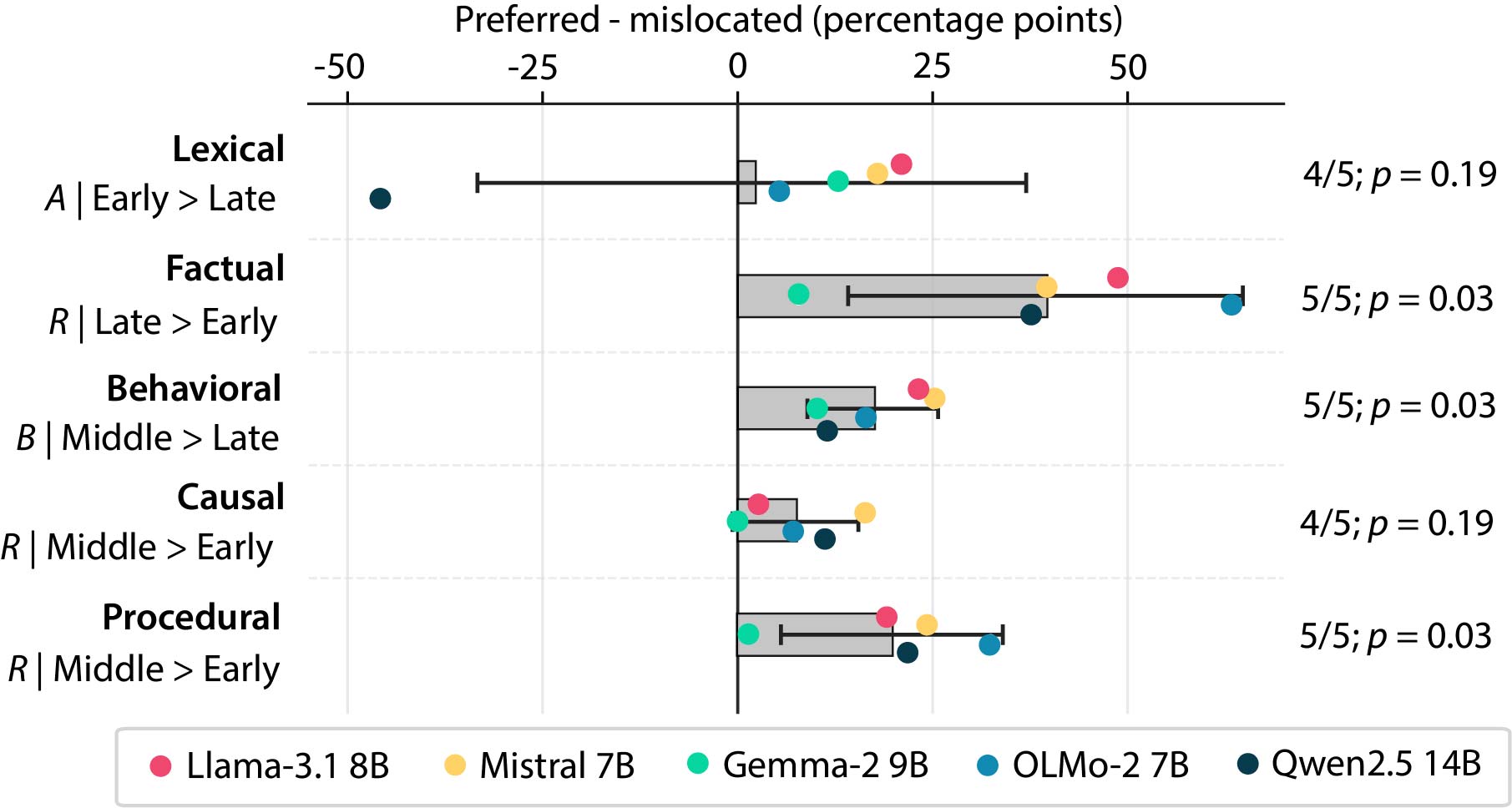}
    \caption{
    Cross-model replication of primary mislocation penalties. Bars show mean preferred vs. mislocated differences across models, error bars show 95\% CIs, and points show individual model means. Positive values indicate replication of the predicted Llama contrast. \(A\), \(R\), and \(B\) denote acquisition, transfer, and boundedness; right-side labels report sign agreement and one-sided exact sign-test \(p\)-values.
    }
    \label{fig:cross-model-mislocation}
\end{figure}

The strongest replication was observed for factual association, behavioral learning, and procedural reasoning (Fig.~\ref{fig:cross-model-mislocation}). Factual association showed the predicted late-over-early transfer advantage in all five models. Behavioral learning replicated the pattern of boundedness in which middle adaptation outperformed late adaptation in all five models, supporting the separation between policy acquisition and gating. Procedural reasoning likewise showed a consistent early-mislocation penalty, with middle outperforming early adaptation on transfer in all five models. Lexical binding and causal mapping were more model-sensitive but still broadly consistent with the primary results. Lexical binding showed an early-over-late acquisition advantage in four of five models, while causal mapping showed a middle-over-early transfer advantage in four of five models. The causal exception was Gemma, whose weak full-stack causal performance under the transferred budget is consistent with model--budget mismatch; a targeted Gemma causal budget-sensitivity analysis in Appendix~E shows substantially improved full-stack acquisition and transfer at larger budgets.

We then tested whether the cross-model geometry profiles contained systematic objective- and model-level structure. For each model, objective, and seed, we represented the localization profile as a localized-minus-full vector across adaptation sites and metrics, as visualized in Appendix~D, Fig.~\ref{fig:cross-model-delta-geometry}. A permutation test over objective labels found significant objective-specific structure in these profiles (observed statistic \(=8.539\), \(p<.001\); 10{,}000 permutations). A corresponding permutation test over model labels also found significant model-specific structure (observed statistic \(=11.679\), \(p<.001\); 10{,}000 permutations). Thus, both learning objective and model family contribute systematically to the observed adaptation geometries.

To compare the relative size of these effects, we computed a descriptive additive sum-of-squares decomposition over the \(75\) model--objective--seed profiles. Objective identity explained \(25.0\%\) of the profile variation, while model identity explained \(34.2\%\). The remaining \(40.8\%\) fell in the residual term, including model--objective interactions and seed-level variation. These results support directional replication of several objective-specific mislocation penalties, but not a strong architecture-invariant account of localization geometry.

Overall, the cross-model analysis strengthens the main result without overstating it. The primary directional contrasts replicate for most objectives across most models, suggesting that adaptation geometry is not an artifact of a single Llama model. At the same time, the significant model-label test, larger model variance component, and Gemma causal sensitivity analysis show that architecture and budget matter. Future work should therefore recalibrate budgets within each model family before drawing stronger claims about universal localization structure.

\section{Discussion}
Across five learning objectives, the regions that best supported acquisition, transfer, and boundedness differed systematically. Lexical binding was most compatible with early-layer adaptation, factual association favored later layers among localized adapters, behavioral policy learning exhibited a distributed profile, and causal mapping and procedural reasoning relied more strongly on middle or full-stack adaptation for transfer. These differences were not reducible to overall task difficulty or adapter parameter count; objective-specific training budgets were calibrated before localization, and the main directional signatures persisted under approximate parameter-matched controls.

\subsection{Adaptation Site is a Functional Design Variable}
These results suggest adaptation site should be treated as a substantive design variable rather than as a mere implementation detail. Full-stack adaptation often provides the strongest raw performance, as expected from its greater number of trainable degrees of freedom. However, localized adaptation reveals distinctions that full-stack performance alone can obscure. Beyond reduced accuracy, poorly matched sites change which component of learning fails. An update may acquire the target response without transferring robustly, transfer without remaining bounded, or appear bounded because it has become overly conservative.

We describe this pattern as an objective's \emph{adaptation geometry}: the profile of how update placement shapes what is learned, how well it generalizes, and how narrowly it is applied. This framing moves beyond treating downstream accuracy as a single scalar outcome by treating placement itself as a source of behavioral variation.

This perspective differs from both model editing and conventional parameter-efficient tuning. Model editing typically evaluates whether a completed update is reliable, generalizes across equivalent prompts, and remains local to its intended scope. Parameter-efficient tuning, by contrast, often chooses adapter placement for computational efficiency or as an approximation to full fine-tuning. Adaptation geometry instead treats placement as an interpretable inductive bias that can alter the behavioral character of the update itself. This view may ultimately support adapters that allocate capacity across depth according to the functional demands of the task.

Importantly, because information is transformed and reused throughout the network, geometry is better understood as a functional intervention map than as evidence of strict modularity. Our results support the claim that adaptation site changes the behavioral profile of an update, but they do not imply that learning objectives correspond to isolated modules or uniquely localized internal representations.

\subsection{Localization as a Diagnostic Tool}

Localization is useful even when a localized adapter is not the best final solution. A small layer-wise sweep can reveal which component of learning a model fails to support. For instance, early adaptation shows that a lexical mapping is easy to acquire but hard to transfer; middle adaptation reveals where bounded policy application is best supported; late adaptation could expose response-level behavior that does not reflect deeper generalization.

This makes localization useful as a diagnostic step before committing to a full adaptation strategy. A coarse sweep may meaningfully indicate whether a task is compatible with a narrowly targeted update, requires broader full-stack intervention, or would benefit from multiple adapters with different roles. In this sense, localized LoRA can also be an experimental probe of how adaptation site shapes learning.

\subsection{Limitations and Future Directions}
This study has several limitations. First, the benchmark is synthetic and isolates five operational learning objectives. While this enables controlled comparison, real adaptation settings often involve overlapping objectives and more complex risks. Future work should evaluate geometry under sequential and large-scale updates, cross-objective interference, adapter removal or reversal, retention on unrelated benchmarks, and safety regressions. Such evaluations would clarify whether localization reduces collateral change or merely redistributes it.

Second, the early, middle, and late windows provide only a coarse view of model depth. Finer-grained analyses of individual layers, overlapping windows, attention and feed-forward modules, and learned mixtures of regions could determine whether the observed profiles reflect broad depth-wise gradients or narrower functional subnetworks.

Third, the study considers only LoRA. Comparisons with full fine-tuning, conventional adapters, prefix-based methods, and targeted model editing could distinguish properties of the underlying model from those induced by the adaptation mechanism. 

Fourth, the cross-model experiments transfer budgets calibrated on Llama rather than independently optimizing them for each model. The results therefore support directional robustness, not architecture-invariant localization. Future work should jointly examine objective, architecture, training duration, and adapter capacity.

These limitations motivate several future directions. One is adaptive placement, where a preliminary localization sweep determines where adapter rank should be allocated. Another is multi-region or compositional adaptation, where separate components are deployed to target acquisition, transfer, and/or boundedness. The present work establishes adaptation geometry as a controlled behavioral diagnostic; future work should examine whether it can also guide more reliable, targeted, and auditable model adaptation.

\clearpage
%\bibliography{references}

% Check whether the conference requires a reproducibility checklist to be included in the paper.
% If so, you can uncomment the following line and ajust the path to include it.
%\input{../../ReproducibilityChecklist/LaTeX/ReproducibilityChecklist.tex}

\clearpage
\section{Appendix A: Tasks and Hypotheses}
\label{app:tasks-hypotheses}

Table~\ref{tab:learning_types} summarizes the five learning objectives used in the benchmark. Each objective is designed to isolate a different kind of learning: attaching a novel lexical form to a concept, learning a new relational fact, applying a context-conditioned behavioral policy, mapping interventions to causal effects, or completing an ordered procedure. For each objective, examples are generated from latent specifications rather than reused templates, allowing us to evaluate the same underlying rule across in-distribution, paraphrased, generalized, and negative-control contexts.

The table also lists the planned localization contrasts used in the main experiments. These contrasts were specified at the level of learning components rather than overall accuracy. For example, lexical binding was predicted to favor early adaptation for acquisition, while factual association was predicted to favor later adaptation for acquisition and transfer. Behavioral policy learning was expected to separate action acquisition from bounded policy application, with late adaptation supporting policy-action selection and middle adaptation supporting policy withholding. Causal mapping and procedural reasoning were predicted to depend more strongly on middle-layer adaptation for transfer, with procedural reasoning also testing whether late-layer boundedness reflects a more conservative profile.

These hypotheses provide the basis for the planned contrasts reported in the main localization results. The examples in Table~\ref{tab:learning_types} are illustrative of the latent-specification format and are intended to show how each objective separates positive, generalization, and negative-control evaluation cases.

\begin{table*}[p]
\centering
\small
\setlength{\tabcolsep}{3pt}
\renewcommand{\arraystretch}{1.18}

\caption{\textbf{Adaptation benchmark suite and planned localization contrasts.}
Each objective targets a distinct computational object. For each objective, we specify the planned directional contrast and a representative latent specification with positive, generalization, and negative-control examples.}
\label{tab:learning_types}

\begin{tabularx}{\linewidth}{@{}
L{0.10\linewidth}
L{0.12\linewidth}
L{0.3\linewidth}
L{0.43\linewidth}
@{}}
\toprule

\textbf{Objective} &
\textbf{Computational Object} &
\textbf{Planned Contrast / Null Hypotheses} &
\textbf{Representative Examples} \\

\midrule

Lexical binding &
Novel form--concept attachment &
$H_1$: Early $>$ Middle/Late on acquisition. \newline
$H_0$: No early-layer advantage. &
\exampleblock
  {``daxel'' means bird.}
  {``What is a daxel?'' $\rightarrow$ bird.}
  {``The daxel built a nest.'' $\rightarrow$ bird.}
  {``Is a daxel a chair?'' $\rightarrow$ no.}
\\

\addlinespace[0.55em]

Factual association &
Subject--relation--object mapping &
$H_1$: Late $>$ Early on acquisition/transfer. \newline
$H_0$: No late-layer advantage. &
\exampleblock
  {Norvessa's capital is Lumo.}
  {``What is the capital of Norvessa?'' $\rightarrow$ Lumo.}
  {``Which city is Norvessa governed from?'' $\rightarrow$ Lumo.}
  {``Is Norvessa's capital Merin?'' $\rightarrow$ no.}
\\

\addlinespace[0.55em]

Behavioral policy learning &
Context-conditioned action selection and withholding &
$H_1$: Late $>$ Middle on acquisition. \newline
$H_1$: Middle $>$ Late on boundedness. \newline
$H_0$: No directional difference. &
\exampleblock
  {Missing email recipient $\rightarrow$ ask for recipient.}
  {``Draft a thank-you email.'' $\rightarrow$ \texttt{ASK\_FOR\_RECIPIENT}.}
  {``Write a follow-up note, but no recipient is named.'' $\rightarrow$ \texttt{ASK\_FOR\_RECIPIENT}.}
  {``Draft a thank-you email to Maya.'' $\rightarrow$ \texttt{NO\_POLICY\_TRIGGER}.}
\\

\addlinespace[0.55em]

Causal mapping &
Directed cause-effect dependency &
$H_1$: Middle $>$ Early/Late on transfer. \newline
$H_0$: No middle-layer advantage. &
\exampleblock
  {Opening Valve A activates cooling and lowers temperature.}
  {``Valve A is opened.'' $\rightarrow$ temperature decreases.}
  {``The operator triggers Valve A during the trial.'' $\rightarrow$ temperature decreases.}
  {``Valve A is blocked before the trial.'' $\rightarrow$ \texttt{NO\_CAUSAL\_EFFECT}.}
\\

\addlinespace[0.55em]

Procedural reasoning &
Ordered step-completion structure &
$H_1$: Middle $>$ Early on transfer. \newline
$H_0$: No middle-layer advantage. \newline
Exploratory: late boundedness vs. acquisition. &
\exampleblock
  {Device is ready only after setup, activation, and verification.}
  {``Setup, activation, and verification are complete.'' $\rightarrow$ ready.}
  {``The device was configured, started, and checked successfully.'' $\rightarrow$ ready.}
  {``Setup and activation are complete, but verification is missing.'' $\rightarrow$ \texttt{PROCEDURE\_INCOMPLETE}.}
\\

\bottomrule
\end{tabularx}
\end{table*}
%\clearpage
%\begin{landscape}

\section{Appendix B: Benchmark Calibration}
%Full-stack calibration confirms which task configurations are learnable under parameter-efficient adaptation and identifies budgets suitable for localization experiments. 

For each learning objective, we construct 25 latent specifications and use the same split structure across objectives. Each specification contributes 12 training examples and 22 evaluation examples: 5 in-distribution (ID) examples, 5 paraphrase examples, 6 generalization examples, and 6 negative-control examples. Thus, for each objective, budget, and seed, ID and paraphrase evaluations contain $n=125$ examples each, while generalization and negative-control evaluations contain $n=150$ examples each. Acquisition is defined as the mean of ID and paraphrase accuracy, and transfer is measured on the generalization split (Gen.). Boundedness is measured on negative controls, but the scoring rule is objective-specific. Specifically, lexical binding and factual association use strict negative-control accuracy, whereas behavioral policy, causal mapping, and procedural reasoning use accuracy on the appropriate null or withholding label.

\begin{figure}[t]
    \centering
    \includegraphics[width=\linewidth]{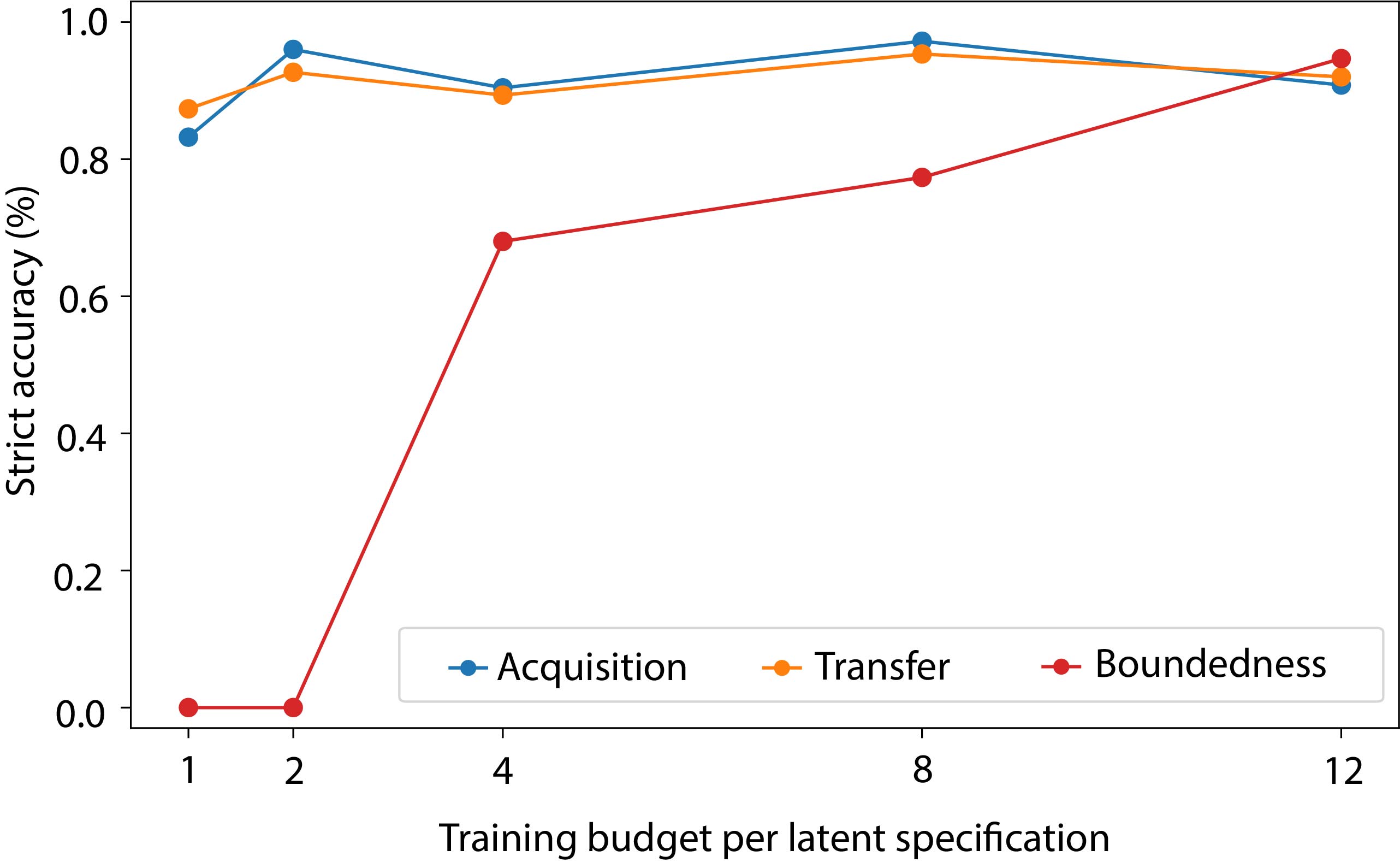}
    \caption{
Full-stack calibration for behavioral learning. Behavioral policy is acquired quickly by budget 2, but boundedness remains at zero until higher budgets. Budget 8 was selected as the provisional calibration point and evaluated further in the neighboring multi-seed sweep shown in Fig.~\ref{fig:behavioral-budget-selection-multiseed}.
}
    \label{fig:behavioral-budget-selection}
\end{figure}

Throughout this appendix, coarse sweep tables report aggregate split accuracy for the initial full-stack budget sweep (Tables~\ref{tab:lexical-binding-calibration-coarse}--\ref{tab:procedural-reasoning-calibration-coarse}). For multi-seed calibration figures, points show mean strict accuracy across seeds, with 95\% bootstrap confidence intervals over seeds and latent specifications (Fig.~\ref{fig:lexical-binding-budget-selection}--\ref{fig:procedural-budget-selection-multiseed}). 

\subsection{Lexical Binding} 

In this task, each latent specification maps a novel surface form, such as a pseudoword or code-like identifier, onto an existing concrete object concept. We evaluate lexical binding in two transfer formats. \emph{Concept transfer} tests whether the model can use the learned word in new contexts and recover the correct underlying concept. \emph{Yes/no transfer} tests whether the model can answer binary questions about the learned mapping with the correct polarity, including both positive and negative formulations. Thus, concept transfer measures whether the mapping has been acquired, while yes/no transfer additionally tests whether the model can apply the mapping reliably across response formats.

\begin{table}[H]
\centering
\small
\setlength{\tabcolsep}{5pt}
\begin{tabular}{lccccc}
\toprule
Condition & Budget & ID & Paraphrase & Gen. & Neg. Ctrl. \\
\midrule
Baseline & -- & 0.0 & 0.0 & 14.0 & 100.0 \\
\midrule
Adapter & 1  & 31.2  & 16.0  & 26.0 & 35.3 \\
Adapter & 2  & 37.6  & 21.6  & 22.7 & 88.7 \\
Adapter & 4  & 72.8  & 64.0  & 47.3 & 61.3 
\\
\rowcolor{black!10} Adapter & 8  & 100.0 & 100.0 & 56.0 & 42.0 \\
Adapter & 12 & 100.0 & 99.2  & 50.0 & 76.0 \\
\bottomrule
\end{tabular}
\caption{
Strict accuracy (\%) in the coarse sweep for lexical binding. Budget 8 is the first coarse budget to reach saturated ID and paraphrase performance, motivating the neighboring multi-seed sweep in Fig.~\ref{fig:lexical-binding-budget-selection}.
}
\label{tab:lexical-binding-calibration-coarse}
\end{table}

Full-stack calibration with Llama-3.1-8B showed that lexical binding was reliably acquired by budget 8 (Table~\ref{tab:lexical-binding-calibration-coarse}). Across three seeds, acquisition increased from $84.1\%$ at budget 6 to $97.7\%$ at budget 8 and $99.7\%$ at budget 10. Concept transfer was already high at budget 6 and increased further from $90.7\%$ to $98.0\%$ and $100.0\%$ across budgets 6, 8, and 10, respectively.

\begin{figure}[H]
    \centering
    \includegraphics[width=\linewidth]{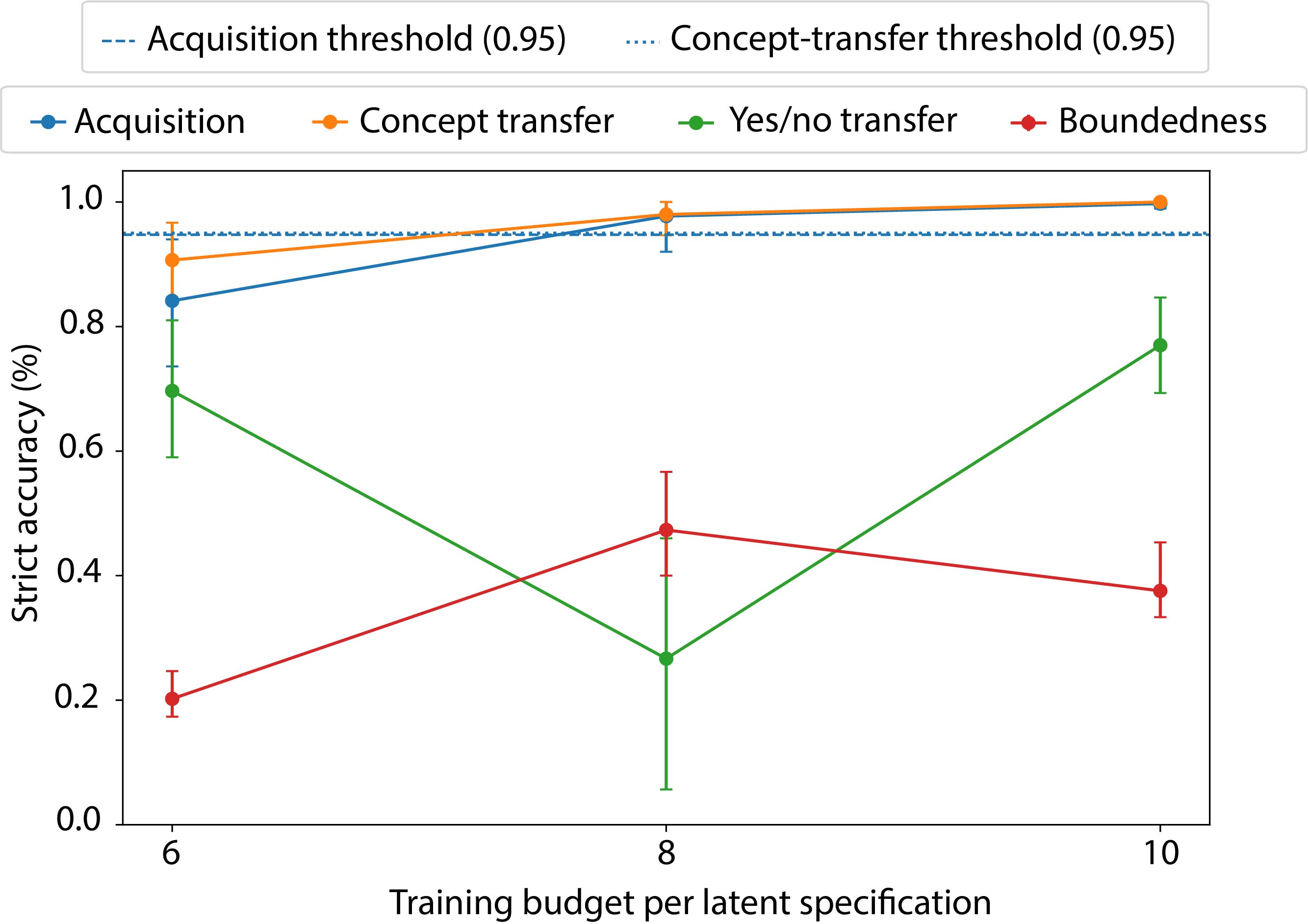}
    \caption{
Multi-seed budget-selection tradeoffs for lexical binding. 
Points show mean performance across seeds; error bars show 95\% CIs. 
Although budget 8 reaches near-saturated acquisition and concept transfer, yes/no transfer drops at this budget, suggesting unstable application across binary response formats. 
Budget 10 preserves saturated acquisition and concept transfer while improving yes/no transfer, and is therefore selected for localization experiments.
}
    \label{fig:lexical-binding-budget-selection}
\end{figure}

We select \textbf{budget 10} as the primary lexical-binding calibration point (Fig.~\ref{fig:lexical-binding-budget-selection}). Although budget 8 is the first budget to achieve near-saturated acquisition and concept transfer, yes/no transfer drops sharply at this budget. This suggests that the model has learned the lexical mapping but is unstable in applying it across binary response formats. Budget 10 preserves saturated acquisition and concept transfer while substantially improving yes/no transfer, making it a cleaner operating point for downstream localization experiments. We report boundedness separately, since negative-control performance does not improve monotonically with budget.

\subsection{Factual Association}
Factual association tests whether the model can learn a new synthetic fact linking two entities through a relation. Each latent specification defines a subject--relation--object association, such as a fictional country and its capital, a person and an invented device, or a person and an affiliated institute. 

\begin{table}[H]
\centering
\small
\setlength{\tabcolsep}{5pt}
\begin{tabular}{lccccc}
\toprule
Condition & Budget & ID & Paraphrase & Gen. & Neg. Ctrl. \\
\midrule
Baseline & -- & 0.0 & 0.0 & 15.3 & 99.3 \\
\midrule
Adapter & 1  & 2.4  & 2.4  & 12.0 & 83.3 \\
Adapter & 2  & 29.6 & 35.2 & 20.7 & 29.3 \\
Adapter & 4  & 56.8 & 60.0 & 48.7 & 30.7 \\
\rowcolor{black!10} Adapter & 8  & 88.8 & 88.8 & 82.0 & 30.7 \\
Adapter & 12 & 90.4 & 90.4 & 78.7 & 31.3 \\
\bottomrule
\end{tabular}
\caption{
Strict accuracy (\%) in the coarse sweep for factual association. 
Full-stack calibration showed a sharp improvement between budgets 4 and 8. 
Budget 8 is the first coarse budget to reach the acquisition--transfer plateau, motivating the neighboring multi-seed sweep in Fig.~\ref{fig:factual-association-budget-selection}. 
%Negative-control performance was high at baseline but dropped after adaptation, indicating that learned associations can spill over to invalid contexts.
}
\label{tab:factual-association-calibration}
\end{table}
\begin{figure}[H]
    \centering
    \includegraphics[width=\linewidth]{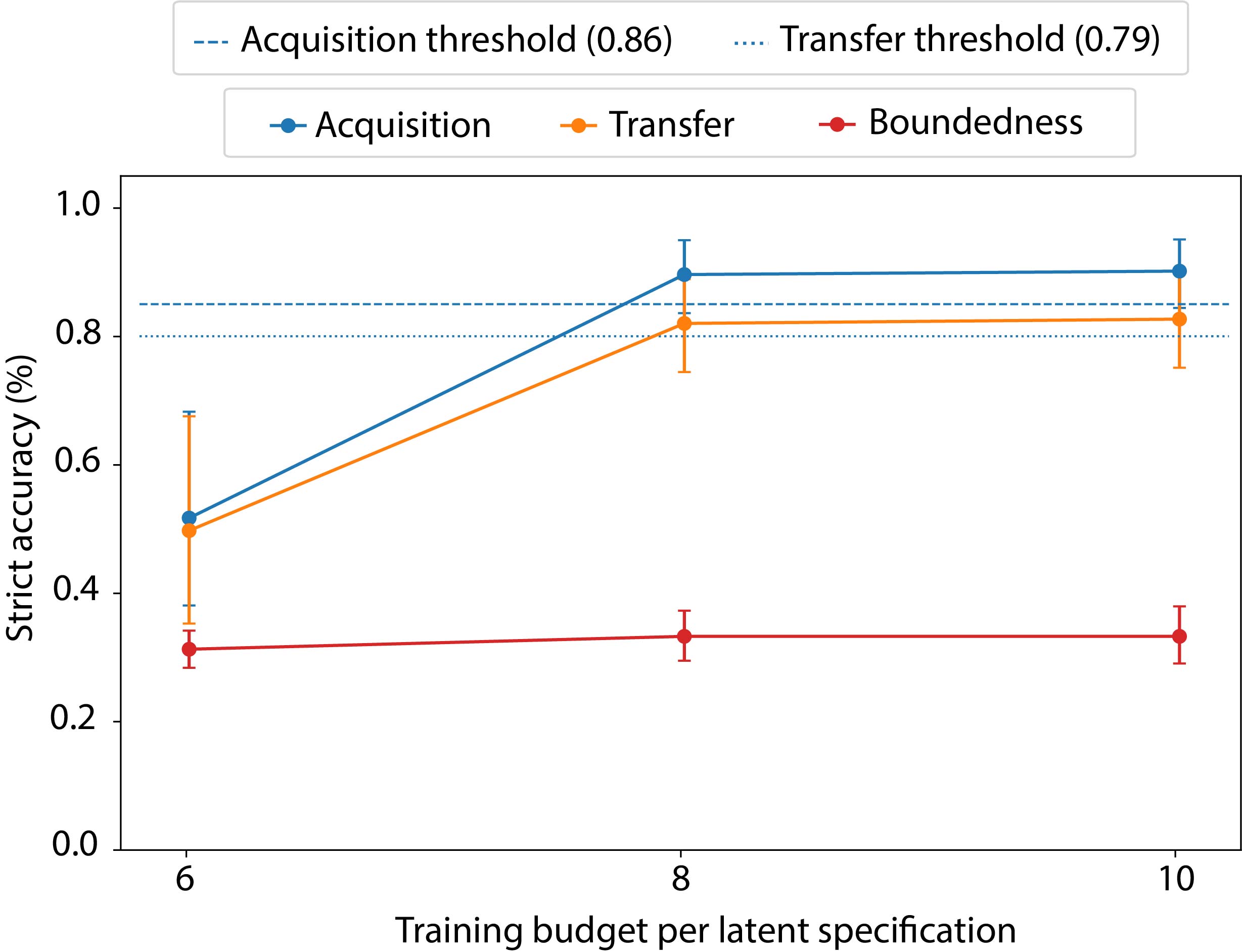}
    \caption{
    Multi-seed budget tradeoffs for factual association. 
    Points show mean strict accuracy across seeds, with 95\% CIs over seeds and latent specifications. 
    Performance increases sharply from budget 6 to budget 8, while budget 10 yields only marginal additional gains. We therefore select budget 8 as the smallest budget on the acquisition--transfer plateau.
    }
    \label{fig:factual-association-budget-selection}
\end{figure}

Full-stack calibration showed that factual association was reliably acquired by budget 8 (Table~\ref{tab:factual-association-calibration}). The high baseline negative-control score is largely because the model has not yet acquired the synthetic fact and therefore cannot over-apply it. Once adaptation succeeds, negative-control performance becomes a stricter test of bounded application---that is, whether the learned relation is recalled only in valid contexts rather than spilling over to nearby but invalid prompts.

Multi-seed calibration showed a sharp transition between budgets 6 and 8 (Fig.~\ref{fig:factual-association-budget-selection}). Acquisition increased from $51.7\%$ at budget 6 to $89.6\%$ at budget 8, with only marginal gain ($90.1\%$) at budget 10. Transfer followed the same pattern, increasing from $49.8\%$ to $82.0\%$ and $82.7\%$ across budgets 6, 8, and 10, respectively. In contrast, negative-control boundedness remained low and nearly flat after adaptation, suggesting that learning the new relation did not necessarily teach the model the scope conditions under which the relation should be withheld. We select \textbf{budget 8} as the primary factual-association calibration point, since it is the earliest budget near the full-stack acquisition and transfer plateau.

\subsection{Behavioral Policy}

Behavioral policy learning tests whether the model can map contextual cues to the appropriate response action. Each latent specification defines a situation--action mapping, such as asking for a missing time, recipient, format, budget, or confirmation before acting. Unlike lexical binding or factual association, successful behavioral policy learning requires not only selecting the correct action when the policy applies, but also withholding the action when the triggering conditions are absent. This \textit{policy-withholding} accuracy is a surrogate measure for boundedness. %Thus, boundedness is measured as \emph{policy-withholding accuracy}, that is, accuracy on negative-control prompts where no learned policy action is warranted.

\begin{table}[H]
\centering
\small
\setlength{\tabcolsep}{5pt}
\begin{tabular}{lccccc}
\toprule
Condition & Budget & ID & Paraphrase & Gen. & Boundedness \\
\midrule
Baseline & -- & 4.0 & 0.0 & 18.7 & 0.0 \\
\midrule
Adapter & 1  & 80.8 & 85.6 & 87.3 & 0.0 \\
Adapter & 2  & 97.6 & 94.4 & 92.7 & 0.0 \\
Adapter & 4  & 91.2 & 89.6 & 89.3 & 68.0 \\
\rowcolor{black!10} Adapter & 8  & 96.8 & 97.6 & 95.3 & 77.3 \\
Adapter & 12 & 92.8 & 88.8 & 92.0 & 94.7 \\
\bottomrule
\end{tabular}
\caption{Strict accuracy (\%) in the coarse sweep for behavioral policy. Boundedness is measured as policy-withholding accuracy on negative controls. Early budgets achieve high acquisition and generalization but fail on boundedness, indicating that learned policies are over-applied when no policy should be triggered.}
\label{tab:behavioral-policy-calibration}
\end{table}

Full-stack calibration showed that policy acquisition can appear strong even at small budgets, but boundedness emerges later (Table~\ref{tab:behavioral-policy-calibration}). In the coarse sweep, budget 1 already reached $83.2\%$ acquisition and $87.3\%$ generalization, and budget 2 reached $96.0\%$ acquisition and $92.7\%$ generalization. However, boundedness remained at $0.0\%$ for both budgets, indicating that the learned policies were over-applied to contexts where no policy should be triggered. Boundedness improved to $68.0\%$ at budget 4, $77.3\%$ at budget 8, and $94.7\%$ at budget 12, suggesting that policy acquisition and policy gating are separable components of adaptation.

\begin{figure}[H]
    \centering
    \includegraphics[width=\linewidth]{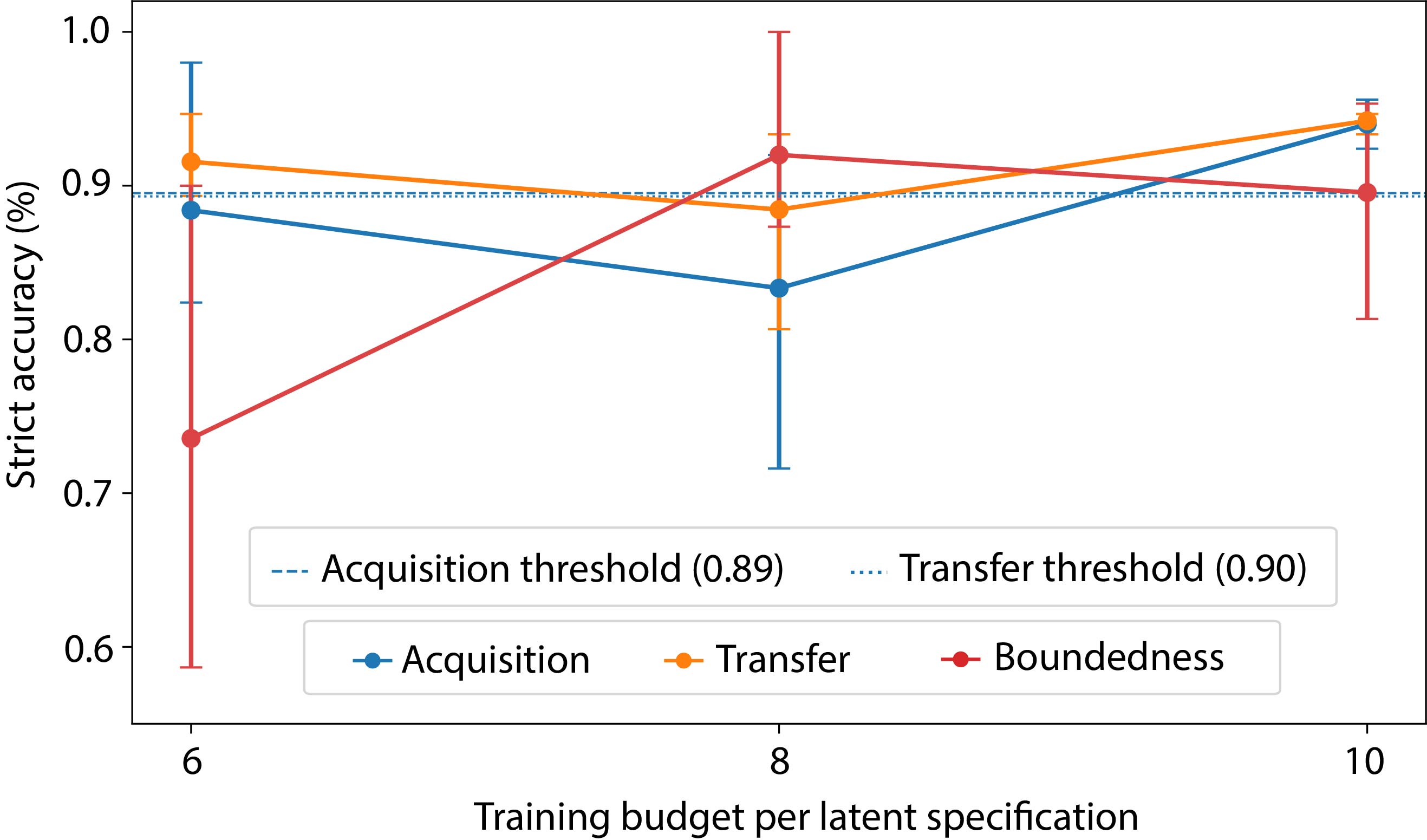}
    \caption{
Multi-seed budget tradeoffs for behavioral policy. Budget 8 achieves strong boundedness but is less stable in acquisition and transfer, while budget 10 provides the most reliable balance across metrics. We therefore select budget 10 as the final calibration point.
}
    \label{fig:behavioral-budget-selection-multiseed}
\end{figure}

The multi-seed sweep (Fig.~\ref{fig:behavioral-budget-selection-multiseed}) further showed that budget 8, while strong in the coarse sweep, was unstable across seeds. Budget 8 achieved high boundedness ($92.0\% \pm 7.0\%$), but lower and more variable acquisition ($83.3\% \pm 10.5\%$) and transfer ($88.4\% \pm 6.8\%$). Budget 10 provided the most reliable balanced operating point, with higher and more stable acquisition ($94.0\% \pm 1.6\%$) and transfer ($94.2\% \pm 0.8\%$), while maintaining strong boundedness ($89.6\% \pm 7.3\%$). We therefore select \textbf{budget 10} as the final calibration point.

\subsection{Causal Mapping}
For causal mapping, each latent specification defines a causal system linking an intervention or activating event to an effect through an underlying mechanism, and the model is trained to map causal prompts to discrete outcome labels. Unlike factual association, where the target is a subject--relation--object mapping, causal mapping requires preserving directionality and applying the rule only when the causal pathway is active. For this objective, boundedness is measured as \emph{null-effect} accuracy---that is, the accuracy on negative-controls where the causal pathway should not produce the learned effect.

\begin{table}[H]
\centering
\small
\setlength{\tabcolsep}{5pt}
\begin{tabular}{lccccc}
\toprule
Condition & Budget & ID & Paraphrase & Gen. & Boundedness \\
\midrule
Baseline & -- & 1.6 & 0.0 & 0.0 & 0.0 \\
\midrule
Adapter & 1  & 34.4  & 21.6  & 32.7 & 0.0 \\
Adapter & 2  & 61.6  & 32.0  & 44.0 & 0.0 \\
Adapter & 4  & 81.6  & 77.6  & 48.0 & 66.7 \\
Adapter & 8  & 97.6  & 92.8  & 76.0 & 66.7 \\
\rowcolor{black!10} Adapter & 12 & 100.0 & 100.0 & 90.0 & 50.0 \\
\bottomrule
\end{tabular}
\caption{
Strict accuracy (\%) in the coarse sweep for causal-mapping calibration.
Full-stack calibration showed rapid improvement in acquisition, while transfer to held-out causal contexts improved more slowly.
Budget 12 produced the strongest coarse transfer, motivating the neighboring multi-seed sweep depicted in Fig.~\ref{fig:causal-mapping-multiseed}.
}
\label{tab:causal-mapping-calibration}
\end{table}

Full-stack calibration showed that direct acquisition and transfer improved at different rates. In the coarse sweep (Table~\ref{tab:causal-mapping-calibration}), acquisition increased from $28.0\%$ at budget 1 to $79.6\%$ at budget 4, $95.2\%$ at budget 8, and $100.0\%$ at budget 12. Transfer improved more slowly, rising from $32.7\%$ at budget 1 to $48.0\%$ at budget 4, $76.0\%$ at budget 8, and $90.0\%$ at budget 12. This separation suggests that the model can learn the direct cause--effect mapping before it can reliably apply the causal rule in held-out contexts.

\begin{figure}[H]
    \centering
    \includegraphics[width=\linewidth]{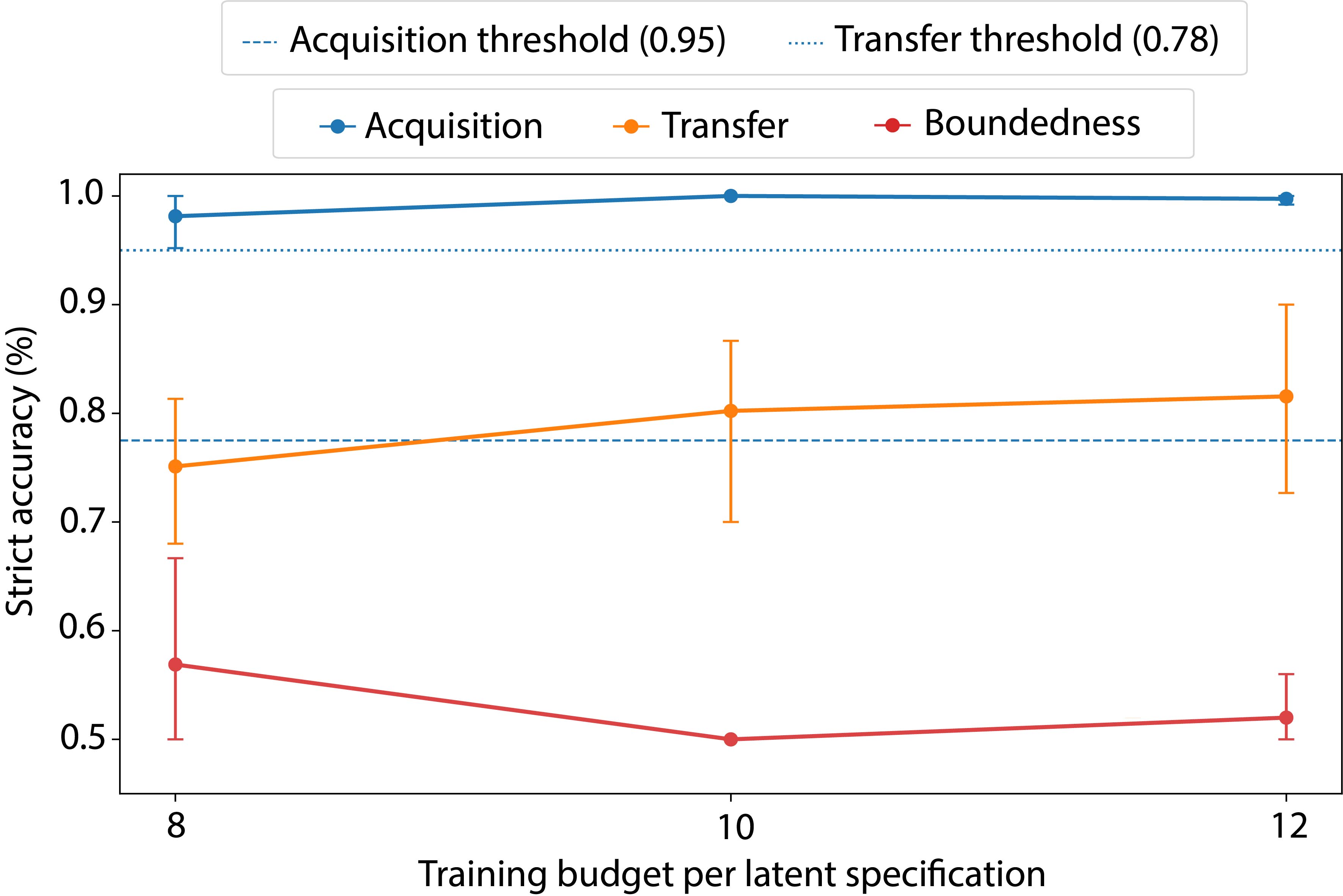}
    \caption{
    Multi-seed budget-selection sweep for causal mapping. Budget 10 reaches saturated acquisition and exceeds the transfer plateau threshold, while budget 12 provides only a small additional transfer gain. 
    We therefore select budget 10 as the final calibration point.
    }
    \label{fig:causal-mapping-multiseed}
\end{figure}

Because the coarse sweep suggested a late transition between budgets 8 and 12, we ran a multi-seed sweep at budgets 8, 10, and 12 (Fig.~\ref{fig:causal-mapping-multiseed}). We did not sweep beyond budget 12 because the goal was to find the smallest budget on the acquisition--transfer plateau; once budgets 10--12 satisfied this criterion, larger budgets would add capacity without changing the calibration decision. 

Budget 8 achieved high acquisition ($98.1\%$) but remained below the transfer threshold ($75.1\%$). Budget 10 reached saturated acquisition ($100.0\%$) and exceeded the transfer threshold ($80.2\%$), while maintaining boundedness comparable to budget 12. Budget 12 produced slightly higher transfer ($81.6\%$), but the gain over budget 10 was small relative to the additional budget. We therefore select \textbf{budget 10} as the final causal-mapping calibration point.

\subsection{Procedural Reasoning}
Procedural reasoning tests whether the model can learn that an outcome depends on completing prerequisite steps in order. Each latent specification defines a procedure with required setup, intermediate actions, and a completion condition. Successful adaptation requires distinguishing valid completions from partial, reordered, or superficially similar sequences. For this objective, boundedness is measured as \emph{incomplete-procedure} accuracy---that is, the accuracy on negative-control prompts where the required procedure has not been validly completed.

\begin{table}[H]
\centering
\small
\setlength{\tabcolsep}{5pt}
\begin{tabular}{lccccc}
\toprule
Condition & Budget & ID & Paraphrase & Gen. & Boundedness \\
\midrule
Baseline & -- & 0.0 & 0.0 & 0.0 & 100.0 \\
\midrule
Adapter & 1  & 31.2  & 0.0  & 0.0  & 26.7 \\
Adapter & 2  & 40.0  & 2.4  & 0.0  & 1.3 \\
Adapter & 4  & 93.6  & 56.8 & 50.0 & 100.0 \\
\rowcolor{black!10} Adapter & 8  & 100.0 & 64.0 & 67.3 & 83.3 \\
Adapter & 12 & 100.0 & 66.4 & 71.3 & 72.0 \\
\bottomrule
\end{tabular}
\caption{
Strict accuracy (\%) in the coarse sweep for procedural-reasoning calibration.
Budget 8 was the first coarse budget to reach saturated ID accuracy with substantial transfer, motivating the neighboring multi-seed sweep.
%Boundedness is measured as incomplete-procedure accuracy on negative controls.
}
\label{tab:procedural-reasoning-calibration-coarse}
\end{table}

Full-stack calibration showed that direct procedural recall emerged earlier than robust procedural transfer. In the coarse sweep (Table~\ref{tab:procedural-reasoning-calibration-coarse}), ID accuracy rose sharply from $31.2\%$ at budget 1 to $93.6\%$ at budget 4 and $100.0\%$ by budget 8. However, paraphrase and transfer improved more gradually in that paraphrase accuracy increased from $56.8\%$ at budget 4 to $64.0\%$ at budget 8 and $66.4\%$ at budget 12, while transfer increased from $50.0\%$ to $67.3\%$ and $71.3\%$ over the same budgets. This suggests that the model can learn the direct procedural label before it can reliably apply the procedure across reworded or held-out contexts.

Boundedness was non-monotonic. It peaked at budget 4, where the model had learned to reject incomplete procedures but had not yet broadly generalized the completion rule. At larger budgets, acquisition and transfer improved, but the learned procedure was applied more aggressively to partial, reordered, or superficially similar sequences, causing incomplete-procedure accuracy to decline. We interpret budget 4 as a bounded but less transferable regime, rather than the best overall calibration point.

%Boundedness peaked at budget 4 because the model had learned to reject incomplete procedures without yet broadly generalizing the completion rule. At larger budgets, acquisition and transfer improved, but the learned procedure was applied more aggressively to partial, reordered, or superficially similar sequences, causing incomplete-procedure accuracy on negative controls to decline. We therefore interpret the peak at budget 4 as a bounded but less transferable regime, rather than the best overall calibration point. Because budget 4 appears only in the coarse sweep, we do not over-interpret this local peak; the multi-seed sweep instead supports the broader conclusion that procedural acquisition and transfer plateau around budgets 8--12 while boundedness does not monotonically improve.

\begin{figure}[H]
    \centering
    \includegraphics[width=\linewidth]{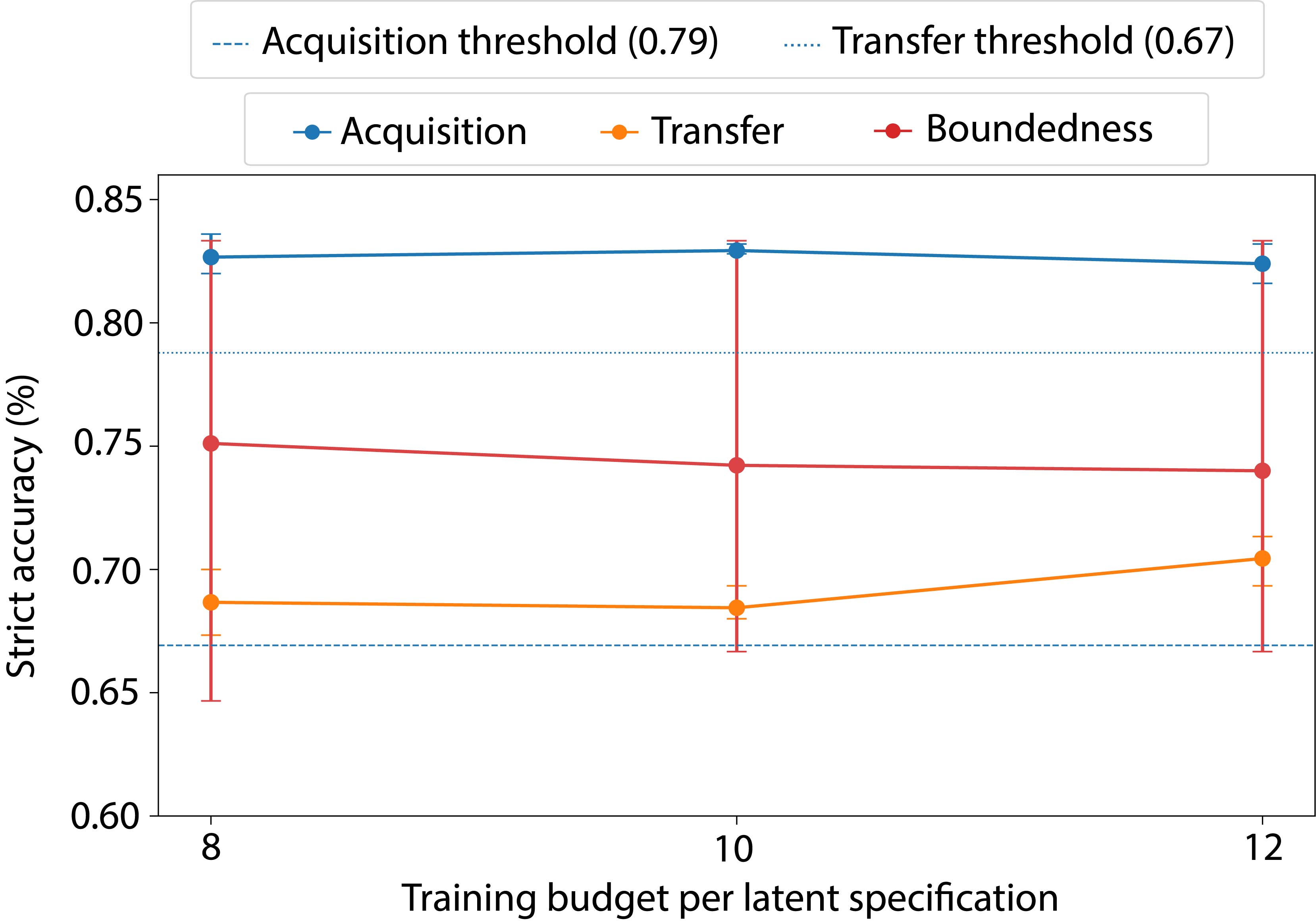}
    \caption{
    Multi-seed budget tradeoffs for procedural reasoning. 
    All candidate budgets fall on a similar acquisition--transfer plateau, but budget 8 is the smallest and has the strongest boundedness among the three. 
    We therefore select budget 8 as the final procedural-reasoning calibration point.
    }
    \label{fig:procedural-budget-selection-multiseed}
\end{figure}

The multi-seed sweep showed that budgets 8, 10, and 12 all lay on a similar acquisition--transfer plateau (Fig.~\ref{fig:procedural-budget-selection-multiseed}). Budget 8 achieved $82.7\%$ acquisition, $68.7\%$ transfer, and the strongest boundedness among the candidate budgets ($75.1\%$). Budget 10 produced nearly identical acquisition and transfer, while budget 12 gave only a marginal transfer increase. We therefore select \textbf{budget 8} as the procedural-reasoning calibration point, since it is the smallest budget on the plateau and preserves the strongest boundedness among the candidates.

\section{Appendix C: Localization Experiments Results}
\label{app:localization}
Table~\ref{tab:cross-objective-localization} presents the full summary for the calibrated objectives under the primary Llama-3.1-8B experiments. %We also report planned seed-level contrasts, an omnibus permutation test, null-label audits, mislocation penalties, and parameter-matched controls. 
\begin{table*}[t]
\centering
\small
\setlength{\tabcolsep}{5pt}
\begin{tabular}{llccc}
\toprule
Objective & Condition & Acquisition & Transfer & Boundedness \\
\midrule
Lexical binding ($B=10$)
  & Full   & \best{99.7}{0.5}   & \best{84.9}{3.3}   & \other{39.7}{6.3} \\
  & Early  & \second{97.2}{2.8} & \other{40.6}{9.3}  & \best{68.0}{7.6} \\
  & Middle & \other{86.4}{1.2}  & \other{50.7}{7.7}   & \second{44.2}{4.3} \\
  & Late   & \other{76.3}{1.3}  & \second{55.0}{1.5} & \other{33.7}{0.3} \\

\addlinespace[0.35em]
Factual association ($B=8$)
  & Full   & \best{89.6}{0.0}   & \best{82.0}{2.7}   & \other{33.3}{2.9} \\
  & Early  & \other{32.4}{2.1}  & \other{20.4}{6.2}  & \other{37.6}{10.8} \\
  & Middle & \other{38.0}{3.5}  & \other{35.3}{2.4}  & \second{46.9}{13.7} \\
  & Late   & \second{52.1}{0.9} & \second{63.1}{0.8} & \best{57.1}{4.4} \\

\addlinespace[0.35em]
Behavioral policy ($B=10$)
  & Full   & \best{94.0}{1.6}   & \best{94.2}{0.8}   & \best{89.6}{7.3} \\
  & Early  & \other{84.1}{4.7}  & \other{88.2}{2.7}  & \other{73.1}{20.1} \\
  & Middle & \other{88.0}{7.3}  & \second{90.4}{8.0} & \second{85.8}{9.7} \\
  & Late   & \second{93.1}{3.0} & \other{87.6}{2.1}  & \other{64.2}{5.2} \\

\addlinespace[0.35em]
Causal mapping ($B=10$)
  & Full   & \best{100.0}{0.0}  & \best{80.2}{9.0}   & \best{50.0}{0.0} \\
  & Early  & \other{93.5}{8.6}  & \other{55.1}{10.7} & \other{28.4}{8.5} \\
  & Middle & \second{98.5}{0.5} & \second{68.0}{1.3} & \best{50.0}{0.0} \\
  & Late   & \other{96.3}{2.4}  & \other{58.7}{2.0}  & \second{44.7}{1.2} \\

\addlinespace[0.35em]
Procedural reasoning ($B=8$)
  & Full   & \best{82.7}{0.8}   & \best{68.7}{1.3}   & \other{75.1}{9.5} \\
  & Early  & \other{74.9}{3.0}  & \other{33.8}{6.0}  & \other{44.0}{8.1} \\
  & Middle & \second{75.6}{4.3} & \second{62.4}{9.6} & \second{76.4}{10.8} \\
  & Late   & \other{60.0}{2.1}  & \other{58.7}{5.8}  & \best{82.0}{0.0} \\
\bottomrule
\end{tabular}
\caption{
Cross-objective localization summary for Llama-3.1-8B. \(B\) denotes the calibrated budget selected for the specific objective. Values are mean accuracy percentages $\pm$ standard deviation in percentage points across seeds $\{11,22,33\}$. Dark-gray cells indicate the best condition within each objective and metric; light-gray cells indicate the second-best distinct condition. %Tied values receive the same designation.
}
\label{tab:cross-objective-localization}
\end{table*}

Rather than testing every pairwise difference, we evaluate a set of planned contrasts derived from each objective's hypothesized localization profile, as specified in Appendix A, Table~\ref{tab:learning_types}. These contrasts are computed from paired seed-level differences. Table~\ref{tab:localization-planned-contrasts} reports the resulting mean differences and bootstrap confidence intervals.

\begin{figure*}[t]
    \centering
    \includegraphics[width=\linewidth]{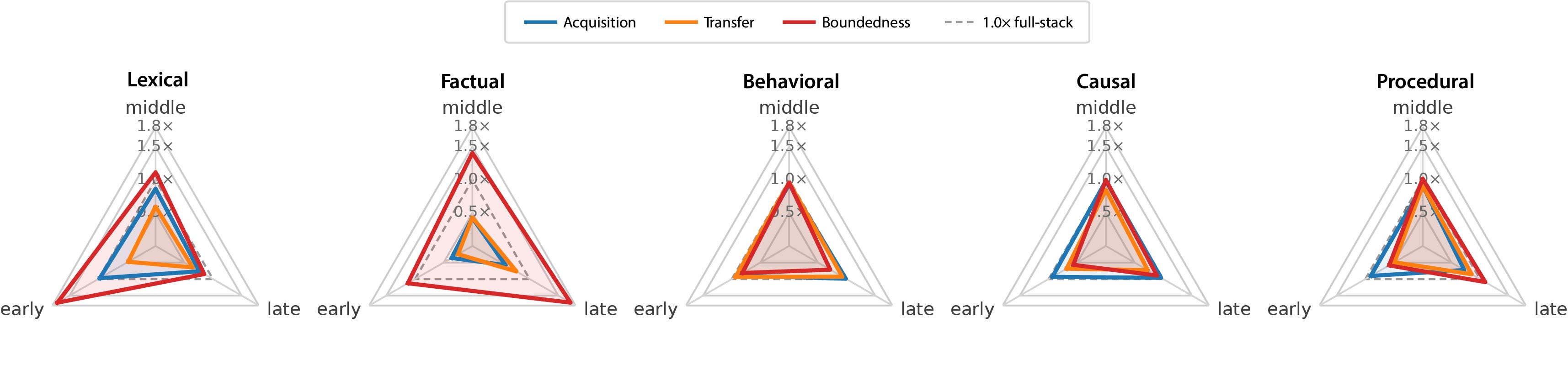}
    \caption{Adaptation geometries in Llama-3.1-8B. Values are normalized by the corresponding full-stack mean for each objective--metric pair; the dashed triangle marks $1.0\times$ parity, and vertices beyond it indicate localized performance above full-stack. Polygon shape captures depth-wise variation, while radial extent shows the proportion of full-stack performance retained or exceeded. The figure recasts Table~\ref{tab:cross-objective-localization} and Fig.~\ref{fig:cross-objective-localization} to highlight how adaptation ``geometries'' deform across learning objectives.}
    \label{fig:adaptation-geometries}
\end{figure*}

Table~\ref{tab:null-label-audits} presents the null-label audits. These analyses show that boundedness is not attributable to global collapse toward the null label in behavioral, causal, or procedural learning, although some localized adapters exhibit more conservative response profiles.

Lastly, Table~\ref{tab:appendix-primary-mislocation-penalties} summarizes the penalties associated with placing each learning objective in a poorly matched adaptation region.

\begin{table*}[!t]
\centering
\small
\begin{tabular}{llr@{\;}c@{\;}lccc}
\toprule
Objective & Metric & \multicolumn{3}{c}{Contrast}
& $\Delta$ (pp) & 95\% CI (pp) & $p$ \\
\midrule
Lexical binding
 & Acquisition & Early  & $>$ & Late   & 20.9 & [17.2, 23.6] & $<.001$ \\
 & Acquisition & Early  & $>$ & Middle & 10.8 & [7.6, 13.2]  & $<.001$ \\
 & Transfer    & Full   & $>$ & Early  & 44.3 & [37.3, 48.0] & $<.001$ \\
 & Boundedness & Early  & $>$ & Full   & 28.3 & [25.0, 30.7] & $<.001$ \\

\addlinespace
Factual association
 & Acquisition & Late   & $>$ & Early  & 19.7 & [18.0, 21.6] & $<.001$ \\
 & Transfer    & Late   & $>$ & Early  & 42.7 & [35.3, 47.3] & $<.001$ \\
 & Transfer    & Full   & $>$ & Late   & 18.9 & [16.7, 22.0] & $<.001$ \\
 & Boundedness & Late   & $>$ & Full   & 23.8 & [22.0, 25.3] & $<.001$ \\

\addlinespace
Behavioral policy
 & Acquisition & Late   & $>$ & Middle & 5.1  & [0.8, 10.4]   & $<.001$ \\
 & Transfer    & Middle & $>$ & Late   & 2.9  & [-5.3, 8.0]   & $.516$ \\
 & Boundedness & Middle & $>$ & Late   & 21.6 & [15.3, 26.7]  & $<.001$ \\
 & Boundedness & Full   & $>$ & Middle & 3.8  & [-1.3, 9.3]   & $.075$ \\

\addlinespace
Causal mapping
 & Transfer    & Middle & $>$ & Early  & 12.9 & [2.0, 22.0]   & $<.001$ \\
 & Transfer    & Middle & $>$ & Late   & 9.3  & [8.7, 10.0]   & $<.001$ \\
 & Transfer    & Full   & $>$ & Middle & 12.2 & [2.0, 17.3]   & $<.001$ \\
 & Boundedness & Middle & $>$ & Early  & 21.6 & [16.7, 31.3]  & $<.001$ \\

\addlinespace
Procedural reasoning
 & Acquisition & Middle & $>$ & Late   & 15.6 & [12.0, 21.2]  & $<.001$ \\
 & Transfer    & Middle & $>$ & Early  & 28.7 & [23.3, 34.0]  & $<.001$ \\
 & Transfer    & Full   & $>$ & Middle & 6.2  & [-1.3, 17.3]  & $.294$ \\
 & Boundedness & Late   & $>$ & Middle & 5.6  & [-1.3, 18.0]  & $.590$ \\

\bottomrule
\end{tabular}
\caption{Planned seed-level contrasts for the Llama-3.1-8B localization experiments. Differences ($\Delta$) and confidence intervals are reported in accuracy percentage points (pp); positive values favor the first condition. Confidence intervals and two-sided bootstrap $p$-values are computed from paired seed-level differences. Given only three seeds, we treat these $p$-values as descriptive and instead emphasize effect sizes and confidence intervals.}
\label{tab:localization-planned-contrasts}
\end{table*}

\begin{table*}[t]
\centering
\small
\setlength{\tabcolsep}{4.5pt}
\renewcommand{\arraystretch}{1.12}

% ---------- High-level summary ----------
\textbf{Aggregate summary}\\[3pt]
\begin{tabular}{llcccc}
\toprule
Objective & Audited null label
& ID ($n=1500$)
& Para. ($n=1500$)
& Gen. ($n=1800$)
& Total ($n=4800$) \\
\midrule

Behavioral policy
& \textsc{no\_policy\_trigger}
& 83 (5.5\%)
& 115 (7.7\%)
& 93 (5.2\%)
& 291 (6.1\%) \\

Causal mapping
& \textsc{no\_causal\_effect}
& ---
& ---
& ---
& --- \\

Procedural reasoning
& \textsc{procedure\_incomplete}
& ---
& 90 (6.0\%)
& 48 (2.7\%)
& 138 (2.9\%) \\

\bottomrule
\end{tabular}

\vspace{0.8em}

% ---------- Condition-level breakdown ----------
\textbf{Breakdown by adaptation condition}\\[3pt]
\begin{tabular}{llcccc}
\toprule
Objective & Condition
& ID ($n=375$)
& Para. ($n=375$)
& Gen. ($n=450$)
& Total ($n=1200$) \\
\midrule

Behavioral policy
& Full
& 13 (3.5\%)
& 13 (3.5\%)
& 22 (4.9\%)
& 48 (4.0\%) \\
& Early
& 34 (9.1\%)
& 29 (7.7\%)
& 18 (4.0\%)
& 81 (6.8\%) \\
& Middle
& 32 (8.5\%)
& 46 (12.3\%)
& 34 (7.6\%)
& 112 (9.3\%) \\
& Late
& 4 (1.1\%)
& 27 (7.2\%)
& 19 (4.2\%)
& 50 (4.2\%) \\

\midrule

Causal mapping
& Full
& ---
& ---
& ---
& --- \\
& Early
& ---
& ---
& ---
& --- \\
& Middle
& ---
& ---
& ---
& --- \\
& Late
& ---
& ---
& ---
& --- \\

\midrule

Procedural reasoning
& Full
& ---
& ---
& ---
& --- \\
& Early
& ---
& ---
& ---
& --- \\
& Middle
& ---
& 11 (2.9\%)
& 27 (6.0\%)
& 38 (3.2\%) \\
& Late
& ---
& 79 (21.1\%)
& 21 (4.7\%)
& 100 (8.3\%) \\

\bottomrule
\end{tabular}

\caption{
Null-label audits for boundedness metrics across the primary
Llama-3.1-8B localization experiments. The upper panel aggregates over
all adaptation conditions; the lower panel reports results separately
for full, early, middle, and late adaptation. Each entry reports how
often the model predicted the audited null label on positive examples
whose correct answer was non-null. For readability, cells containing ``---'' represent $0$ ($0.0\%$). Low rates indicate that boundedness
is not explained by global collapse to the null label, while
condition-level differences reveal more conservative profiles for some
localized adapters.
}
\label{tab:null-label-audits}
\end{table*}

\begin{table*}[!t]
\centering
\small
\setlength{\tabcolsep}{2.8pt}
\renewcommand{\arraystretch}{1.12}
\begin{tabular}{ll r@{\;}l r@{\;}l r@{\;}l r@{\;}l}
\toprule
& & \multicolumn{2}{c}{Acquisition} & \multicolumn{2}{c}{Transfer} & \multicolumn{2}{c}{Boundedness} & \multicolumn{2}{c}{Balanced mean} \\
\cmidrule(lr){3-4}
\cmidrule(lr){5-6}
\cmidrule(lr){7-8}
\cmidrule(lr){9-10}
Objective & Contrast
& $\Delta$ & 95\% CI
& $\Delta$ & 95\% CI
& $\Delta$ & 95\% CI
& $\Delta$ & 95\% CI \\
\midrule

Lexical binding
& Early $-$ Late
& $20.9$ & $[17.2, 23.6]$
& $-14.4$ & $[-19.3, -5.3]$
& $34.3$ & $[29.3, 42.7]$
& $13.6$ & $[11.2, 15.9]$ \\

Factual association
& Late $-$ Early
& $19.7$ & $[18.0, 21.6]$
& $42.7$ & $[35.3, 47.3]$
& $19.6$ & $[3.3, 30.7]$
& $27.3$ & $[23.4, 30.5]$ \\

Behavioral policy
& Middle $-$ Late
& $-5.1$ & $[-10.4, -0.8]$
& $2.9$ & $[-5.3, 8.0]$
& $21.6$ & $[15.3, 26.7]$
& $6.5$ & $[3.6, 8.2]$ \\

Causal mapping
& Middle $-$ Early
& $5.1$ & $[-1.2, 15.2]$
& $12.9$ & $[2.0, 22.0]$
& $21.6$ & $[16.7, 31.3]$
& $13.2$ & $[6.6, 20.4]$ \\

Procedural reasoning
& Middle $-$ Early
& $0.7$ & $[-4.0, 7.6]$
& $28.7$ & $[23.3, 34.0]$
& $32.4$ & $[27.3, 39.3]$
& $20.6$ & $[17.5, 25.2]$ \\

\bottomrule
\end{tabular}
\caption{
Primary mislocation penalties for Llama-3.1-8B across learning objectives. Values are percentage-point differences between the preferred localized condition and the least-preferred or most diagnostic mislocated condition, with 95\% bootstrap confidence intervals. Positive values indicate that the preferred adaptation site improves performance on that metric. Balanced mean is the average of acquisition, transfer, and boundedness.
}
\label{tab:appendix-primary-mislocation-penalties}
\end{table*}

\subsection{Parameter-Matched Controls}
\label{app:parameter-matched-controls}
Because the main experiments match rank per adapted layer, full-stack adapters have greater total capacity than localized adapters. We therefore either increased localized rank or reduced full-stack rank to match their rank–layer products, testing whether localization signatures persist when approximate adapter capacity is controlled for.

Table~\ref{tab:parameter-matched-controls-app} summarizes these parameter-matched results. Across both controls, the main localization signatures are largely preserved. Lexical binding remains strongest in early layers for acquisition and boundedness. Factual association remains late-localized among constrained adapters. Behavioral policy learning retains its distributed profile, with late adaptation best for action acquisition and middle adaptation best for transfer and boundedness. Causal mapping continues to favor middle adaptation for transfer and boundedness. Procedural reasoning continues to favor middle adaptation for transfer, while late adaptation remains the most bounded localized condition. Thus, the main adaptation geometries are not explained by trainable parameter count alone.

We also audited whether boundedness in these controls could be explained by trivial collapse to the null label. Table~\ref{tab:param-matched-null-label-audits} reports null-label predictions on positive examples for behavioral policy, causal mapping, and procedural reasoning. 
%Causal mapping showed no collapse to \textsc{no\_causal\_effect} under either control. Behavioral policy showed modest rates of \textsc{no\_policy\_trigger} predictions on positive examples, but far below what would be expected under global policy-withholding collapse. Procedural reasoning likewise showed no global collapse to \textsc{procedure\_incomplete}, although late-layer adapters were more conservative, especially on paraphrase examples. 
These audits support the interpretation that the parameter-matched boundedness results are not artifacts of uniformly defaulting to the null label, while also preserving the finding that late-layer adaptation for procedural reasoning trades stronger boundedness for a more conservative profile.

% Requires:
% \usepackage{booktabs}
% \usepackage{multirow}

\begin{table*}[!t]
\centering
\small
\setlength{\tabcolsep}{4.5pt}
\renewcommand{\arraystretch}{1.08}

\begin{tabular}{lllcccc}
\toprule
Objective
& Capacity control
& Condition
& Rank $\times$ layers
& Acquisition
& Transfer
& Boundedness \\
\midrule

Lexical binding
& Expanded localized
& Full
& $8 \times 32 = 256$
& $1.000 \pm 0.000$
& $0.764 \pm 0.125$
& $0.433 \pm 0.047$ \\
&
& Early
& $32 \times 8 = 256$
& $0.944 \pm 0.052$
& $0.511 \pm 0.112$
& $0.587 \pm 0.092$ \\
&
& Middle
& $32 \times 8 = 256$
& $0.863 \pm 0.032$
& $0.598 \pm 0.251$
& $0.411 \pm 0.084$ \\
&
& Late
& $32 \times 8 = 256$
& $0.761 \pm 0.014$
& $0.533 \pm 0.035$
& $0.342 \pm 0.010$ \\
\cmidrule(lr){2-7}
& Reduced full-stack
& Full
& $2 \times 32 = 64$
& $0.999 \pm 0.002$
& $0.529 \pm 0.048$
& $0.511 \pm 0.120$ \\
&
& Early
& $8 \times 8 = 64$
& $0.975 \pm 0.031$
& $0.642 \pm 0.222$
& $0.464 \pm 0.057$ \\
&
& Middle
& $8 \times 8 = 64$
& $0.872 \pm 0.028$
& $0.544 \pm 0.136$
& $0.429 \pm 0.068$ \\
&
& Late
& $8 \times 8 = 64$
& $0.767 \pm 0.021$
& $0.549 \pm 0.008$
& $0.342 \pm 0.008$ \\

\midrule

Factual association
& Expanded localized
& Full
& $8 \times 32 = 256$
& $0.883 \pm 0.017$
& $0.829 \pm 0.054$
& $0.331 \pm 0.017$ \\
&
& Early
& $32 \times 8 = 256$
& $0.353 \pm 0.016$
& $0.189 \pm 0.050$
& $0.369 \pm 0.062$ \\
&
& Middle
& $32 \times 8 = 256$
& $0.376 \pm 0.004$
& $0.353 \pm 0.044$
& $0.464 \pm 0.158$ \\
&
& Late
& $32 \times 8 = 256$
& $0.501 \pm 0.019$
& $0.624 \pm 0.008$
& $0.560 \pm 0.059$ \\
\cmidrule(lr){2-7}
& Reduced full-stack
& Full
& $2 \times 32 = 64$
& $0.872 \pm 0.021$
& $0.860 \pm 0.012$
& $0.409 \pm 0.143$ \\
&
& Early
& $8 \times 8 = 64$
& $0.352 \pm 0.056$
& $0.164 \pm 0.047$
& $0.364 \pm 0.053$ \\
&
& Middle
& $8 \times 8 = 64$
& $0.404 \pm 0.028$
& $0.336 \pm 0.020$
& $0.418 \pm 0.066$ \\
&
& Late
& $8 \times 8 = 64$
& $0.509 \pm 0.012$
& $0.651 \pm 0.025$
& $0.573 \pm 0.082$ \\

\midrule

Behavioral policy
& Expanded localized
& Full
& $8 \times 32 = 256$
& $0.929 \pm 0.008$
& $0.929 \pm 0.034$
& $0.911 \pm 0.017$ \\
&
& Early
& $32 \times 8 = 256$
& $0.753 \pm 0.045$
& $0.829 \pm 0.042$
& $0.831 \pm 0.023$ \\
&
& Middle
& $32 \times 8 = 256$
& $0.892 \pm 0.042$
& $0.927 \pm 0.023$
& $0.887 \pm 0.012$ \\
&
& Late
& $32 \times 8 = 256$
& $0.923 \pm 0.037$
& $0.853 \pm 0.018$
& $0.649 \pm 0.074$ \\
\cmidrule(lr){2-7}
& Reduced full-stack
& Full
& $2 \times 32 = 64$
& $0.913 \pm 0.032$
& $0.920 \pm 0.029$
& $0.936 \pm 0.037$ \\
&
& Early
& $8 \times 8 = 64$
& $0.803 \pm 0.013$
& $0.873 \pm 0.029$
& $0.860 \pm 0.064$ \\
&
& Middle
& $8 \times 8 = 64$
& $0.903 \pm 0.025$
& $0.920 \pm 0.031$
& $0.931 \pm 0.010$ \\
&
& Late
& $8 \times 8 = 64$
& $0.948 \pm 0.008$
& $0.880 \pm 0.012$
& $0.682 \pm 0.071$ \\

\midrule

Causal mapping
& Expanded localized
& Full
& $8 \times 32 = 256$
& $1.000 \pm 0.000$
& $0.758 \pm 0.033$
& $0.440 \pm 0.093$ \\
&
& Early
& $32 \times 8 = 256$
& $0.993 \pm 0.006$
& $0.613 \pm 0.105$
& $0.344 \pm 0.019$ \\
&
& Middle
& $32 \times 8 = 256$
& $0.985 \pm 0.005$
& $0.640 \pm 0.069$
& $0.500 \pm 0.000$ \\
&
& Late
& $32 \times 8 = 256$
& $0.937 \pm 0.021$
& $0.611 \pm 0.057$
& $0.447 \pm 0.012$ \\
\cmidrule(lr){2-7}
& Reduced full-stack
& Full
& $2 \times 32 = 64$
& $0.989 \pm 0.018$
& $0.733 \pm 0.111$
& $0.500 \pm 0.000$ \\
&
& Early
& $8 \times 8 = 64$
& $0.979 \pm 0.008$
& $0.629 \pm 0.083$
& $0.389 \pm 0.096$ \\
&
& Middle
& $8 \times 8 = 64$
& $0.983 \pm 0.005$
& $0.656 \pm 0.043$
& $0.500 \pm 0.000$ \\
&
& Late
& $8 \times 8 = 64$
& $0.959 \pm 0.027$
& $0.616 \pm 0.004$
& $0.473 \pm 0.033$ \\

\midrule

Procedural reasoning
& Expanded localized
& Full
& $8 \times 32 = 256$
& $0.824 \pm 0.017$
& $0.644 \pm 0.102$
& $0.849 \pm 0.104$ \\
&
& Early
& $32 \times 8 = 256$
& $0.724 \pm 0.079$
& $0.444 \pm 0.068$
& $0.691 \pm 0.217$ \\
&
& Middle
& $32 \times 8 = 256$
& $0.784 \pm 0.024$
& $0.609 \pm 0.094$
& $0.693 \pm 0.159$ \\
&
& Late
& $32 \times 8 = 256$
& $0.560 \pm 0.026$
& $0.509 \pm 0.037$
& $0.822 \pm 0.004$ \\
\cmidrule(lr){2-7}
& Reduced full-stack
& Full
& $2 \times 32 = 64$
& $0.817 \pm 0.006$
& $0.698 \pm 0.014$
& $0.867 \pm 0.047$ \\
&
& Early
& $8 \times 8 = 64$
& $0.780 \pm 0.011$
& $0.431 \pm 0.085$
& $0.111 \pm 0.071$ \\
&
& Middle
& $8 \times 8 = 64$
& $0.775 \pm 0.061$
& $0.653 \pm 0.035$
& $0.804 \pm 0.027$ \\
&
& Late
& $8 \times 8 = 64$
& $0.597 \pm 0.023$
& $0.553 \pm 0.029$
& $0.824 \pm 0.004$ \\

\bottomrule
\end{tabular}

\caption{
Parameter-matched controls for the primary localization experiments.
Expanded-localized controls increase the rank of localized adapters to
match the capacity of full-stack adaptation, whereas reduced-full-stack
controls decrease the full-stack rank to match the capacity of localized
adaptation. Adapter capacity is approximated as LoRA rank multiplied by
the number of adapted layers, as shown in the Rank $\times$ layers
column. Values report means $\pm$ standard deviations across three seeds.
}
\label{tab:parameter-matched-controls-app}
\end{table*}

\begin{table*}[!t]
\centering
\small
\setlength{\tabcolsep}{4.2pt}
\renewcommand{\arraystretch}{1.12}

% ---------- High-level summary ----------
\textbf{Aggregate summary}\\[3pt]
\begin{tabular}{lllcccc}
\toprule
Objective & Control & Audited null label
& ID ($n=1500$)
& Para. ($n=1500$)
& Gen. ($n=1800$)
& Total ($n=4800$) \\
\midrule

Behavioral policy
& Expanded
& \textsc{no\_policy\_trigger}
& 111 (7.4\%)
& 155 (10.3\%)
& 109 (6.1\%)
& 375 (7.8\%) \\

& Reduced
& \textsc{no\_policy\_trigger}
& 97 (6.5\%)
& 146 (9.7\%)
& 100 (5.6\%)
& 343 (7.1\%) \\

\midrule

Causal mapping
& Expanded
& \textsc{no\_causal\_effect}
& ---
& ---
& ---
& --- \\

& Reduced
& \textsc{no\_causal\_effect}
& ---
& ---
& ---
& --- \\

\midrule

Procedural reasoning
& Expanded
& \textsc{procedure\_incomplete}
& ---
& 127 (8.5\%)
& 82 (4.6\%)
& 209 (4.4\%) \\

& Reduced
& \textsc{procedure\_incomplete}
& ---
& 100 (6.7\%)
& 48 (2.7\%)
& 148 (3.1\%) \\

\bottomrule
\end{tabular}

\vspace{0.8em}

% ---------- Condition-level breakdown ----------
\textbf{Breakdown by adaptation condition}\\[3pt]
\begin{tabular}{lllcccc}
\toprule
Objective & Control & Condition
& ID ($n=375$)
& Para. ($n=375$)
& Gen. ($n=450$)
& Total ($n=1200$) \\
\midrule

Behavioral policy
& Expanded
& Full
& 12 (3.2\%)
& 19 (5.1\%)
& 21 (4.7\%)
& 52 (4.3\%) \\
& & Early
& 66 (17.6\%)
& 58 (15.5\%)
& 36 (8.0\%)
& 160 (13.3\%) \\
& & Middle
& 26 (6.9\%)
& 46 (12.3\%)
& 27 (6.0\%)
& 99 (8.2\%) \\
& & Late
& 7 (1.9\%)
& 32 (8.5\%)
& 25 (5.6\%)
& 64 (5.3\%) \\

\cmidrule(lr){2-7}
& Reduced
& Full
& 20 (5.3\%)
& 34 (9.1\%)
& 23 (5.1\%)
& 77 (6.4\%) \\
& & Early
& 55 (14.7\%)
& 45 (12.0\%)
& 36 (8.0\%)
& 136 (11.3\%) \\
& & Middle
& 20 (5.3\%)
& 42 (11.2\%)
& 27 (6.0\%)
& 89 (7.4\%) \\
& & Late
& 2 (0.5\%)
& 25 (6.7\%)
& 14 (3.1\%)
& 41 (3.4\%) \\

\midrule

Causal mapping
& Expanded
& Full
& ---
& ---
& ---
& --- \\
& & Early
& ---
& ---
& ---
& --- \\
& & Middle
& ---
& ---
& ---
& --- \\
& & Late
& ---
& ---
& ---
& --- \\

\cmidrule(lr){2-7}
& Reduced
& Full
& ---
& ---
& ---
& --- \\
& & Early
& ---
& ---
& ---
& --- \\
& & Middle
& ---
& ---
& ---
& --- \\
& & Late
& ---
& ---
& ---
& --- \\

\midrule

Procedural reasoning
& Expanded
& Full
& ---
& ---
& ---
& --- \\
& & Early
& ---
& ---
& ---
& --- \\
& & Middle
& ---
& 10 (2.7\%)
& 26 (5.8\%)
& 36 (3.0\%) \\
& & Late
& ---
& 117 (31.2\%)
& 56 (12.4\%)
& 173 (14.4\%) \\

\cmidrule(lr){2-7}
& Reduced
& Full
& ---
& 2 (0.5\%)
& ---
& 2 (0.2\%) \\
& & Early
& ---
& ---
& ---
& --- \\
& & Middle
& ---
& ---
& 2 (0.4\%)
& 2 (0.2\%) \\
& & Late
& ---
& 98 (26.1\%)
& 46 (10.2\%)
& 144 (12.0\%) \\

\bottomrule
\end{tabular}

\caption{
Null-label audits for boundedness metrics under the parameter-matched controls. The upper panel aggregates over all adaptation conditions; the lower panel reports results separately for full, early, middle, and late adaptation. Each entry reports how often the model predicted the audited null label on positive examples whose correct answer was non-null. ``Expanded'' denotes the localized-expanded-rank control, and ``Reduced'' denotes the full-reduced-rank control. For readability, cells containing ``---'' represent $0$ ($0.0\%$). Low rates indicate that boundedness under the parameter-matched controls is not explained by global collapse to the null label, while condition-level differences reveal more conservative profiles for some localized adapters.
}
\label{tab:param-matched-null-label-audits}
\end{table*}

\section{Appendix D: Cross-Model Robustness}
\label{app:cross-model-robustness}
\begin{figure*}[p]
    \centering
    \includegraphics[width=0.9\linewidth]{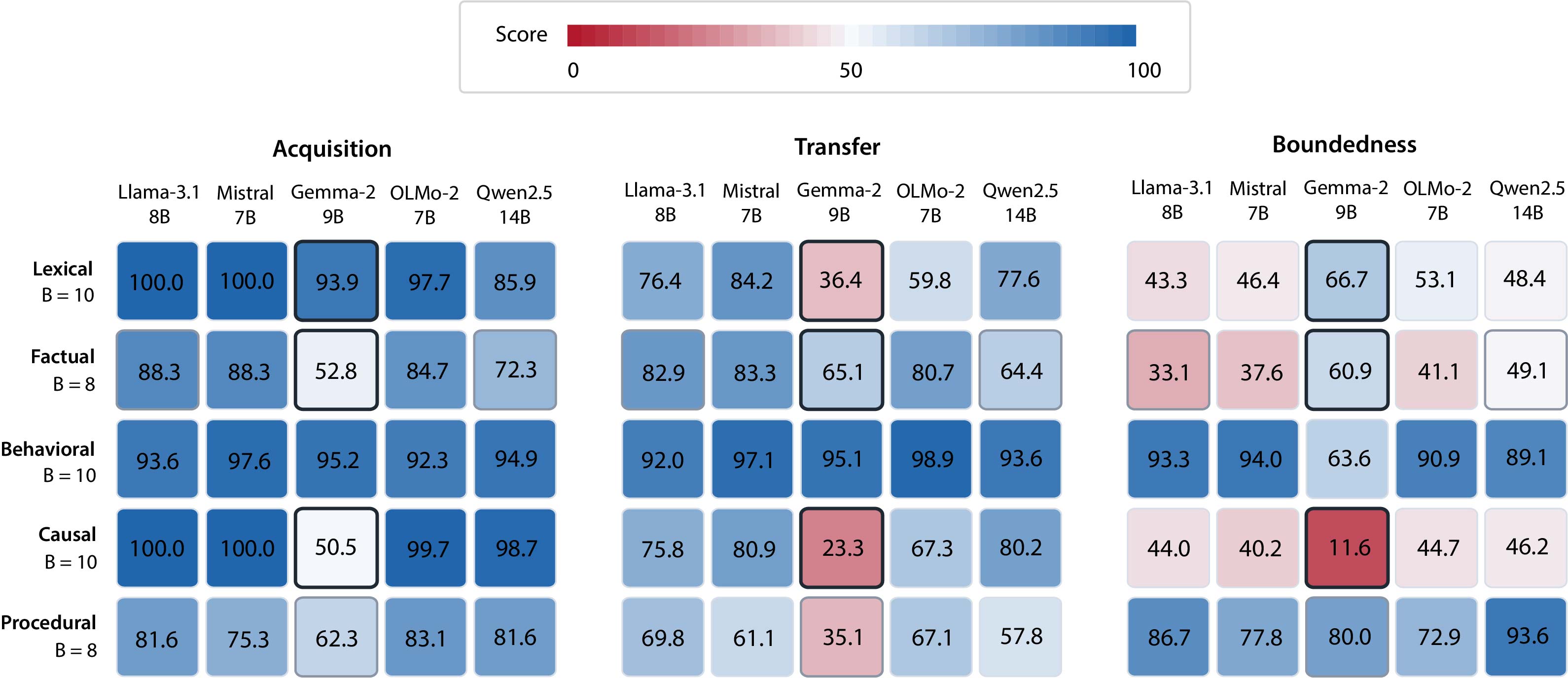}
    \caption{
Full-stack performance under transferred budgets. 
Rows correspond to learning objectives, with \(B\) denoting the objective-specific adaptation budget selected on Llama-3.1-8B and then reused for all models in that row. 
Cell values report mean full-stack performance across three seeds (\%).
Color encodes score magnitude, with red indicating lower performance and blue indicating higher performance. 
Bold outlines mark model--objective pairs flagged by a screening rule as possible model--budget mismatches: acquisition below \(60\%\), transfer below \(35\%\), or performance at least \(25\) pp lower than Llama on acquisition or \(35\) pp lower than Llama on transfer. This diagnostic supports the interpretation that weak cross-model localization profiles should be treated cautiously when full-stack performance is low under the transferred budget.
}
    \label{fig:cross-model-full-stack-performance}
\end{figure*}

Fig.~\ref{fig:cross-model-full-stack-performance} reports full-stack performance at the transferred budgets across all five models. This diagnostic identifies cases where a model may not have learned the objective well under the Llama-selected budget. Therefore, weak full-stack acquisition or transfer should be treated as possible model--budget mismatch rather than definitive evidence about that model's adaptation geometry.

\begin{figure*}[p]
    \centering
    \includegraphics[width=0.75\linewidth]{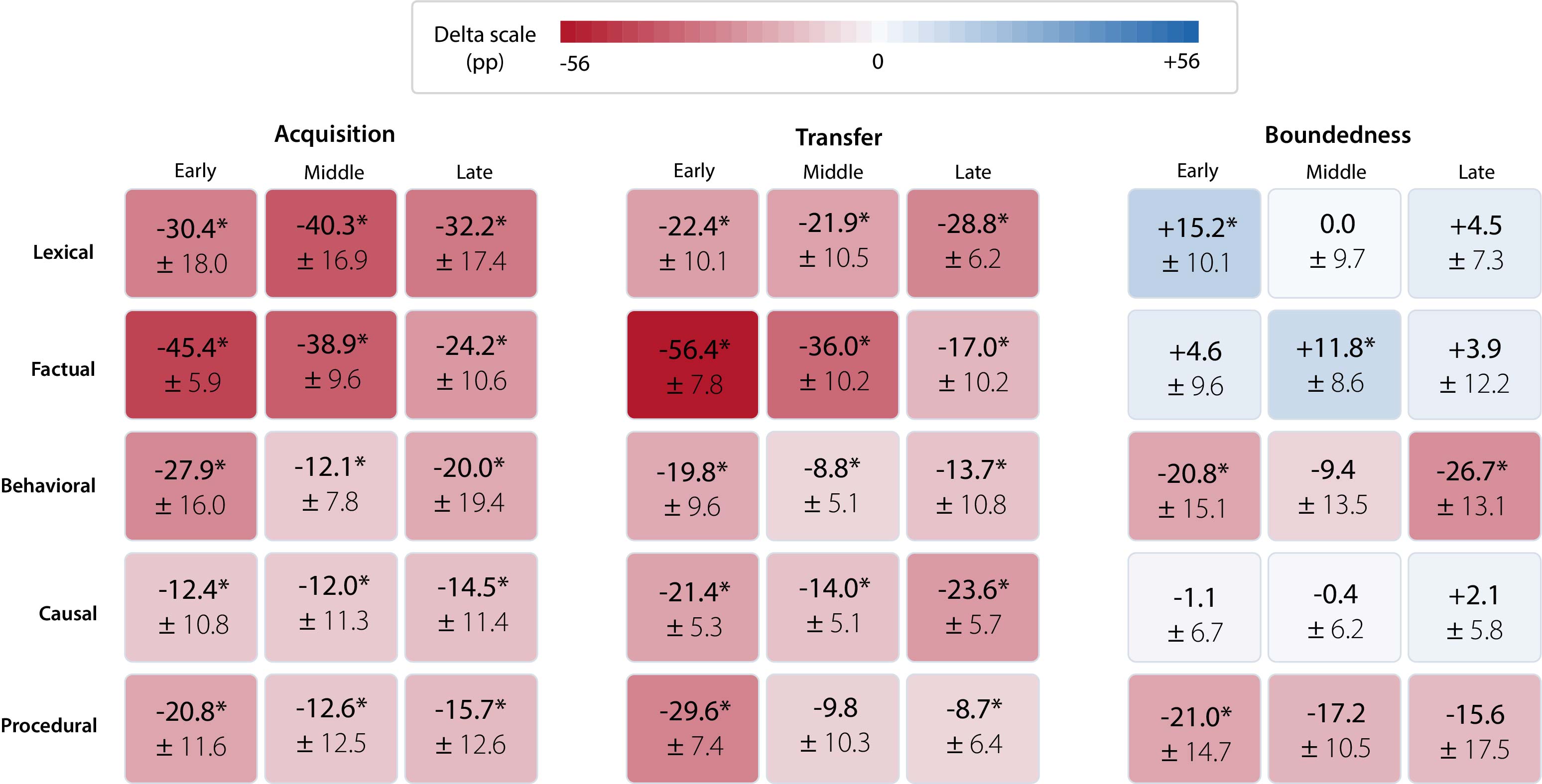}
    \caption{
    Cross-model localized-minus-full adaptation geometry under the transferred-budget protocol. 
    Each cell reports the mean paired delta, in percentage points, between a localized adapter and full-stack adaptation for the same model, objective, and seed (\(n=15\) model--seed pairs per cell). 
    Negative values indicate localized underperformance relative to full-stack adaptation; positive values indicate localized improvement. 
    Cell color encodes the signed delta, the second line gives the 95\% confidence-interval half-width, and asterisks mark intervals excluding zero. Expecting full-stack to be the empirical upper bound, this shows how much each localized region differs by objective and metric.
    }
    \label{fig:cross-model-delta-geometry}
\end{figure*}

Table~\ref{tab:cross-model-directional-replication}, depicted in the main text as Fig.~\ref{fig:cross-model-mislocation}, summarize the primary directional mislocation contrasts across models. Fig.~\ref{fig:cross-model-delta-geometry} provides a complementary full-stack-reference view by comparing each localized adapter directly against full-stack adaptation. The pattern of differences vary across objectives, metrics, and adaptation sites. This supports the same broader interpretation as the primary mislocation analysis, depicted in Fig. \ref{fig:cross-model-mislocation}, that cross-model localization is not governed by a single universal depth preference, but by objective-specific tradeoffs that remain partly model-dependent.

Overall, the cross-model results support directional robustness without implying architecture-invariant localization. The primary mislocation penalties replicate most strongly for factual association, behavioral policy learning, and procedural reasoning, while lexical binding and causal mapping show greater sensitivity to the model--budget pair.

\begin{table*}[t]
\centering
\small
\setlength{\tabcolsep}{4.2pt}
\renewcommand{\arraystretch}{1.12}
\begin{tabular}{lllrrrrrcc}
\toprule
Objective & Contrast & Metric
& Gemma & Llama & Mistral & OLMo & Qwen 
& Replication & $p$ \\
\midrule

Lexical binding
& Early $>$ Late & Acquisition
& $12.9$ & $20.8$ & $17.9$ & $5.2$ & $-47.7$
& 4/5 & .188 \\

Factual association
& Late $>$ Early & Transfer
& $7.8$ & $48.7$ & $39.8$ & $63.3$ & $37.8$
& 5/5 & .031 \\

Behavioral policy
& Middle $>$ Late & Boundedness
& $10.0$ & $23.3$ & $25.1$ & $16.4$ & $11.8$
& 5/5 & .031 \\

Causal mapping
& Middle $>$ Early & Transfer
& $0.0$ & $2.7$ & $16.2$ & $7.1$ & $11.1$
& 4/5 & .188 \\

Procedural reasoning
& Middle $>$ Early & Transfer
& $1.3$ & $19.1$ & $24.2$ & $32.4$ & $21.8$
& 5/5 & .031 \\

\bottomrule
\end{tabular}
\caption{
Cross-model replication of directional contrasts under transferred-budgets. Model columns report percentage-point differences between the preferred and mislocated adaptation regions, averaged across three seeds. Positive values indicate that the contrast follows the predicted direction from the primary Llama experiment. Replication counts report the number of models with a positive contrast. $p$ is the one-sided exact sign-test \(p\)-value under \(H_0 = 0.5\).
}
\label{tab:cross-model-directional-replication}
\end{table*}

\section{Appendix E: Budget Sensitivity Analysis}
\label{app:gemma-causal-sensitivity}

Because Gemma showed the weakest causal-mapping performance under the transferred-budget protocol (shown relative to the other four models in Table~\ref{tab:cross-model-directional-replication}), we ran a targeted full-stack budget sensitivity analysis for Gemma-2-9B causal mapping. This analysis explores whether weak Gemma causal localization under the Llama-selected budget could reflect insufficient adaptation budget.

\begin{figure}[H]
    \centering
    \includegraphics[width=\linewidth]{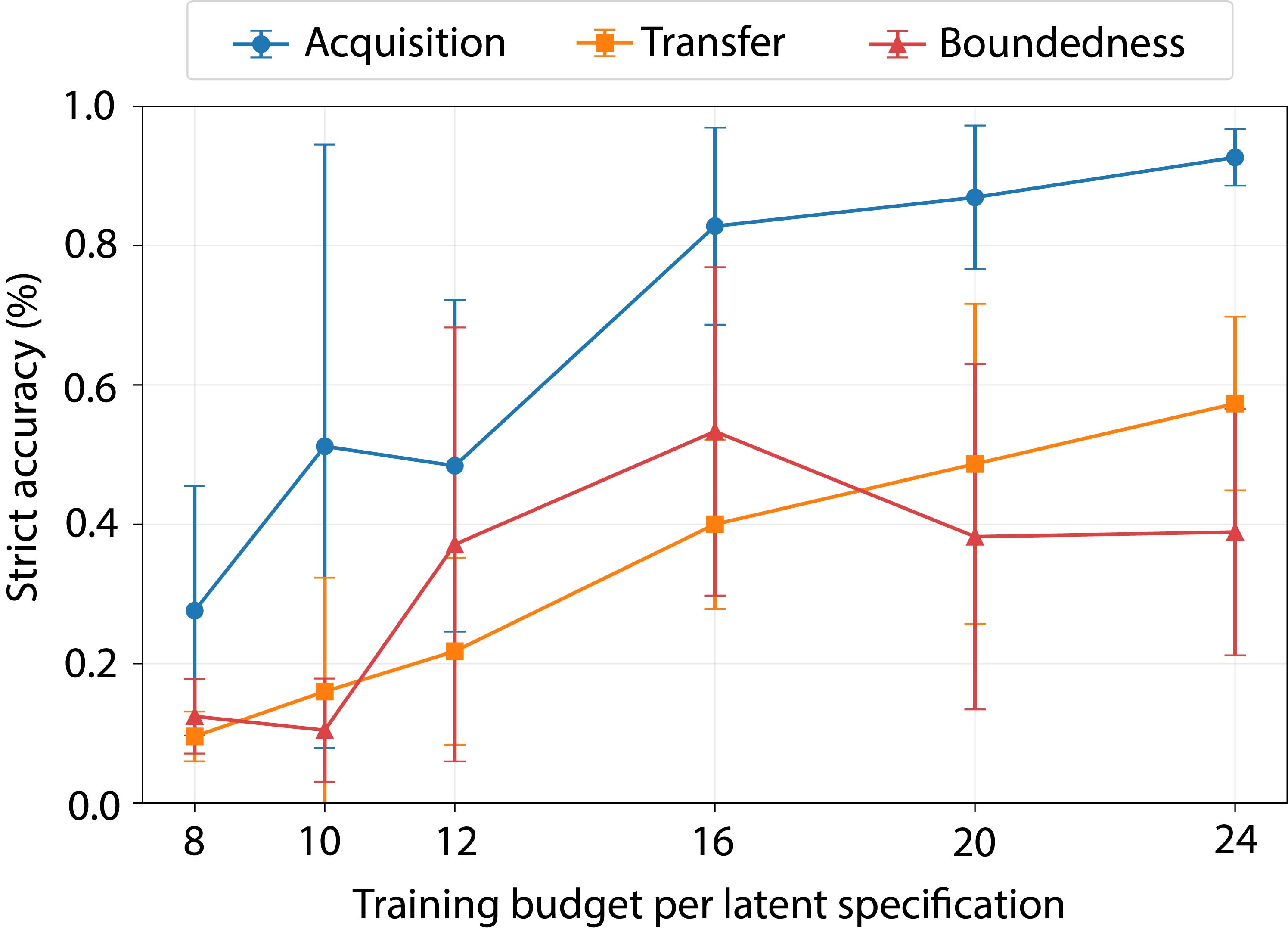}
    \caption{
    Gemma causal-mapping budget sensitivity under full-stack adaptation.
    Points show mean strict accuracy across seeds and error bars show 95\% CIs. Gemma improves at larger budgets, indicating that weak causal-mapping performance under the transferred Llama budget (B = 10) is consistent with model--budget mismatch rather than clear evidence that the objective is not learnable for this model.
    }
    \label{fig:gemma-causal-budget-sensitivity}
\end{figure}

Fig.~\ref{fig:gemma-causal-budget-sensitivity} shows that Gemma causal performance improves substantially with larger budgets. At the transferred budget ($B = 10$), full-stack adaptation reached only \(51.2\%\) acquisition and \(16.0\%\) transfer. Increasing the budget to \(B=24\) raised acquisition to \(92.7\%\) and transfer to \(57.3\%\). This supports the interpretation that Gemma's weak causal profile reflects model--budget mismatch rather than definitive evidence against a causal adaptation geometry. %Because this sensitivity analysis was full-stack only, we retain the original transferred-budget results for the cross-model comparison and treat model-specific recalibration as future work.

\begin{comment}
\begin{table}[!t]
\centering
\small
\setlength{\tabcolsep}{5pt}
\renewcommand{\arraystretch}{1.12}
\begin{tabular}{cccc}
\toprule
Budget & Acquisition & Transfer & Boundedness \\
\midrule
8  & $27.6 \pm 15.8$ & $9.6 \pm 3.2$   & $12.4 \pm 4.7$ \\
10 & $51.2 \pm 38.3$ & $16.0 \pm 14.4$ & $10.4 \pm 6.5$ \\
12 & $48.4 \pm 21.0$ & $21.8 \pm 11.9$ & $37.1 \pm 27.5$ \\
16 & $82.8 \pm 12.5$ & $40.0 \pm 10.7$ & $53.3 \pm 20.8$ \\
20 & $86.9 \pm 9.1$  & $48.7 \pm 20.3$ & $38.2 \pm 21.9$ \\
24 & $92.7 \pm 3.6$  & $57.3 \pm 11.0$ & $38.9 \pm 15.6$ \\
\bottomrule
\end{tabular}
\caption{
Gemma causal-mapping full-stack budget sensitivity. Values are mean percentages \(\pm\) standard deviations across three seeds. Performance improves substantially beyond the transferred Llama-selected budget, indicating that weak Gemma causal performance under the cross-model protocol is consistent with model--budget mismatch.
}
\label{tab:gemma-causal-budget-sensitivity}
\end{table}
\end{comment}

\end{document}